\documentclass[10pt,twocolumn,letterpaper]{article}

\usepackage[pagenumbers]{cvpr} 
\usepackage{multirow}
\usepackage{colortbl}
\usepackage{xcolor}
\usepackage{bbm}
\usepackage{makecell}
\usepackage{booktabs}

\definecolor{cvprblue}{rgb}{0.21,0.49,0.74}
\usepackage[pagebackref,breaklinks,colorlinks,citecolor=cvprblue]{hyperref}

\def\paperID{4692} 
\def\confName{CVPR}
\def\confYear{2024}

\title{AvatarGPT: All-in-One Framework for Motion Understanding, Planning, Generation and Beyond }
\author{Zixiang Zhou\\
{\tt\small zhouzixiang@xiaobing.ai}
\and
Yu Wan\\
{\tt\small wanyu@xiaobing.ai}
\and
Baoyuan Wang\\
{\tt\small wangbaoyuan@xiaobing.ai}
}

\begin{document}

\twocolumn[{
\maketitle
\vspace{-35pt}
\begin{center}
    \includegraphics[width=1.0\textwidth]{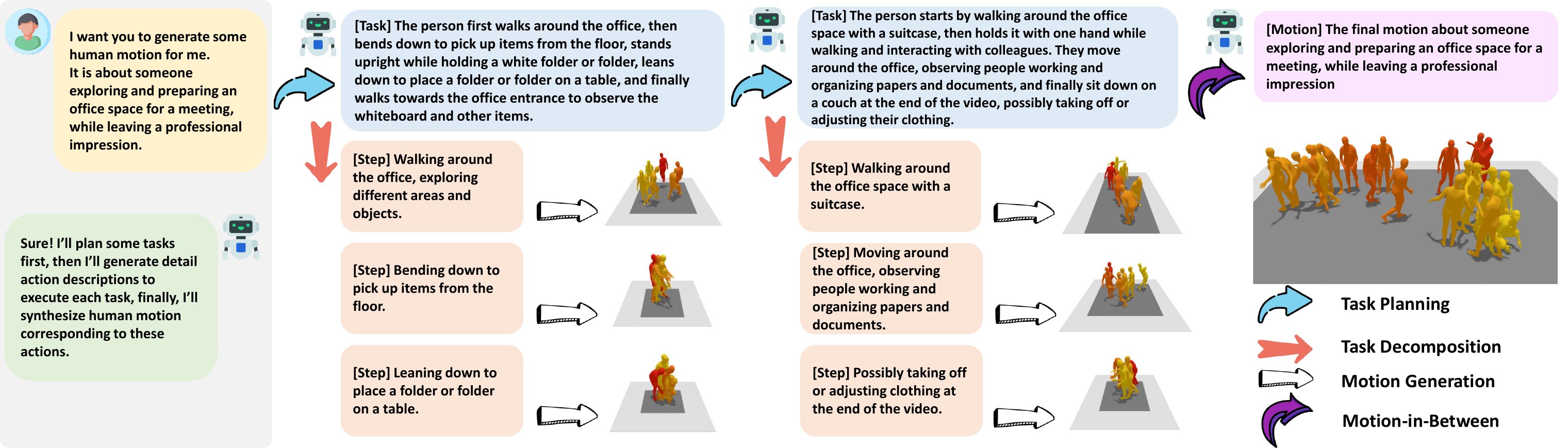}
    \captionof{figure}{An example of long human motion generation based on high-level user instructions, powered by the traversal of a few key modules within our proposed framework, including motion task planning, decomposition, generation, and motion in-between synthesis. }
    \label{fig:teaser}
\end{center}
}]

\vspace{-4pt}
\maketitle

\begin{abstract}
\vspace{-8pt}
Large Language Models(LLMs) have shown remarkable emergent abilities in unifying almost all (if not every) NLP tasks. In the human motion-related realm, however, researchers still develop siloed models for each task. Inspired by InstuctGPT\cite{instructGPT22}, and the generalist concept behind Gato \cite{reed2022generalist}, we introduce \textbf{AvatarGPT}, an All-in-One framework for motion understanding, planning, generations as well as other tasks such as motion in-between synthesis. AvatarGPT treats each task as one type of instruction fine-tuned on the shared LLM. All the tasks are seamlessly interconnected with language as the universal interface, constituting a closed-loop within the framework. To achieve this, human motion sequences are first encoded as discrete tokens, which serve as the extended vocabulary of LLM. Then, an unsupervised pipeline to generate natural language descriptions of human action sequences from in-the-wild videos is developed. Finally, all tasks are jointly trained. Extensive experiments show that AvatarGPT achieves SOTA on low-level tasks, and promising results on high-level tasks, demonstrating the effectiveness of our proposed All-in-One framework. Moreover, for the first time, AvatarGPT enables a principled approach by iterative traversal of the tasks within the closed-loop for unlimited long-motion synthesis. Project website: \url{https://zixiangzhou916.github.io/AvatarGPT/}
\end{abstract}

\section{Introduction}
Text-based human motion generation has made significant progress in recent years\cite{petrovich2022temos,petrovich2023tmr,tevet2022human,zhou2023ude,guo2022tm2t,zhang2023remodiffuse,chen2023executing}. These varied methods fundamentally aim to learn a direct mapping from natural language descriptions to human motions. Despite their impressive generative performance, they exhibit two limitations when viewed from an end-to-end perspective.  Firstly, their efficacy predominantly lies in generating short motion sequences, and extending these methods to longer durations presents substantial challenges. Secondly, the dependency on manually crafted, detailed textual inputs for each motion sequence constrains their utility in real-world scenarios. Consider, for instance, the context of video game environments where non-player characters (NPCs) are required to perform multifaceted, high-level tasks, which need to be broken down into smaller, sequentially ordered sub-tasks, each demanding specific textual descriptions to guide the associated movements. Additionally, the need for post-processing to ensure seamless transitions between motions further complicates the application. Therefore, there is a pressing need to develop a comprehensive end-to-end framework. This framework should not only automate the process of motion planning but also proficiently handle task decomposition and motion generation, all while understanding the contextual subtleties and conforming to broad user-defined instructions.

Fortunately, we have witnessed significant advancement of Large Language Models(LLMs) as well as their remarkable emergent abilities in text-centric understanding\cite{wei2022finetuned,flann2,instructGPT22}, reasoning \cite{xiao2023unified,park2023generative,sun2023adaplanner,huang2022language,song2023llm} and generations \cite{brown2020language,chung2022scaling,ouyang2022training,radford2018improving,radford2019language,raffel2020exploring,touvron2023llama,touvron2023llama2}. Works like InstructGPT \cite{instructGPT22} and Gato\cite{reed2022generalist} further push the paradigm-shifting to a new stage where almost all tasks can be treated as different instructions fine-tuned on top of the shared foundation LLM. However, in the realm of human body motion, these tasks continue to be addressed in isolation. There has been limited exploration in integrating these emergent LLM capabilities to innovate and validate new approaches in this domain.

\begin{figure*}[h]
    \centering
    \includegraphics[width=\linewidth]{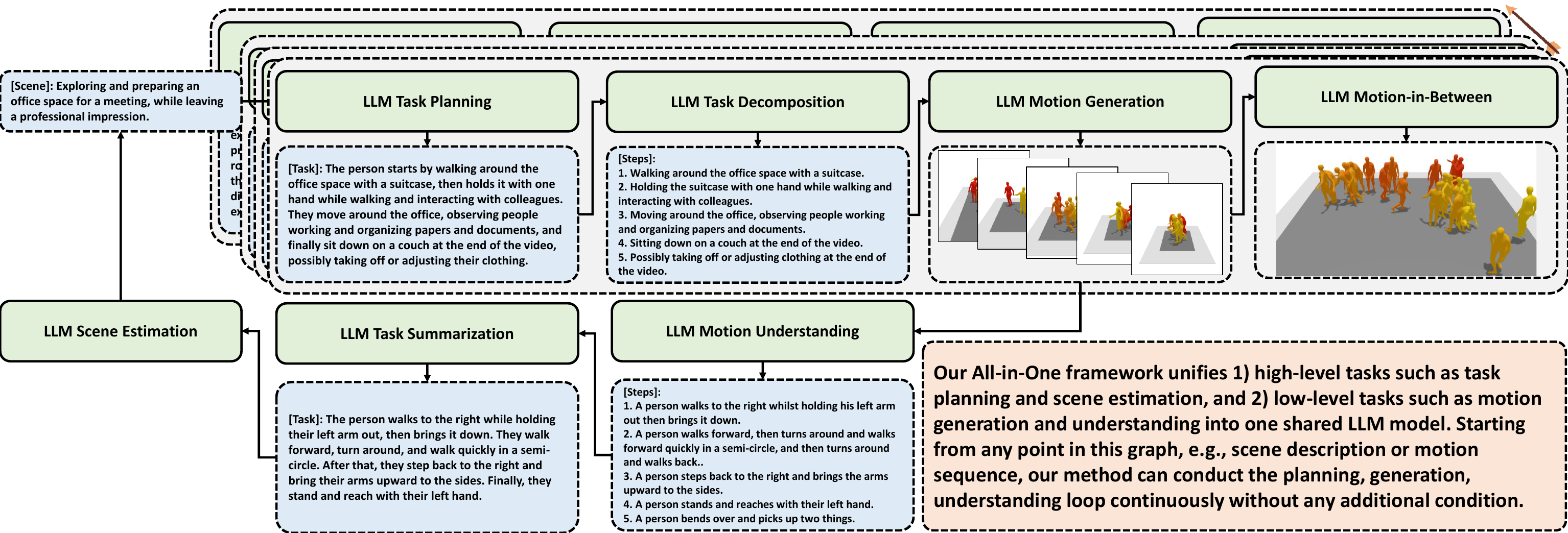}
    \caption{Overview of our All-in-One framework. Language serves as the interface to connect all these modulated tasks within the loop.}
    \label{fig:pipeline-1}
\end{figure*}

Given the context, building on these paradigm shifts, we present \textbf{AvatarGPT}, a groundbreaking unified framework that harnesses the power of LLMs for seven distinct motion-related tasks, including motion understanding, planning, decomposition, generation, motion in-between synthesis, and scene estimation as well as task summarization. Note that, motion planning is a high-level task that requires a deep understanding of the context as well as certain common sense and reasoning abilities. The overview design is illustrated in Fig.\ref{fig:pipeline-1}. We represent continuous motion sequences as discrete tokens. To integrate motion modality into LLMs, we treat the discrete motion tokens as an extended vocabulary, so it could be integrated into any LLMs with little effort. AvatarGPT is structured around four high-level sub-tasks: task planning, decomposition, summarization, and scene estimation, as well as three low-level sub-tasks: text-driven motion generation, motion understanding, and in-between motion generation. We employ specialized prompts for each sub-task and use an instruction-tuning strategy for model training. Additionally, we have developed an innovative method to construct a dataset from real-world videos for high-level instruction tuning, which notably eliminates the need for manual human intervention in the process. To sum up, our contributions are as follows:
\begin{itemize}
    \item We pioneer an All-in-One framework that integrates both high-level and low-level motion-related tasks, fostering a comprehensive optimization loop across understanding, planning, and generation phases.
    \item We develop a novel pipeline to construct a dataset from in-the-wild videos and also curate a dataset specifically for fine-tuning high-level human action planning.
    \item Through extensive evaluation, we demonstrate that our method sets new state-of-the-art benchmarks in low-level tasks and shows promising results in high-level tasks.
    \item Our framework significantly extends the capability for longer synthesis of human motions compared to prior works, thus paving the way for new applications.
\end{itemize}

\section{Related Work}

\paragraph{Motion Generation}
The synthesis of realistic human motion from natural language descriptions is a longstanding research area. Initial efforts like \cite{guo2020action2motion} and \cite{petrovich2021action} generated motion based on pre-defined action categories rather than natural language, limiting the diversity and control of generated motions. Recent advances \cite{tevet2022human, zhou2023ude, petrovich2022temos, petrovich2023tmr, guo2022tm2t, chen2023executing, zhang2023remodiffuse, zhang2023t2m} have addressed this by using natural language descriptions as direct inputs for motion generation. These methods vary in their technical approaches: (1) VAE-based methods, such as \cite{petrovich2022temos} and \cite{petrovich2023tmr}, focus on learning an aligned space between motion and language, utilizing a decoder for motion synthesis. (2) Diffusion models, used in \cite{chen2023executing} and \cite{tevet2022human}, learn to guide diffusion processes for generating human motion distributions from Gaussian noise. Additional auxiliary conditions can be applied for more detailed control \cite{yuan2023physdiff, karunratanakul2023guided}. (3) Tokenization, a widespread technique in language models, has recently been applied to human motion generation \cite{zhou2023ude, zhang2023t2m, guo2022tm2t, jiang2023motiongpt, zhang2023motiongpt}, treating the process as akin to predicting the next motion token in language models. Despite showing promise, these methods still rely on specific user inputs, which can limit their broader applicability.\par

\paragraph{Motion Understanding}
Understanding the meaning of human motion is also a long-term research topic. Describing human motion with pre-defined action labels \cite{zhu2023motionbert}\cite{chi2022infogcn} has dominated this topic for a while. However, these methods have obvious drawbacks, they are not appropriate to describe complex motion sequences. Recently, the learning of the relationship between motion sequence and natural language description has attracted increasing attention. For instance, \cite{plappert2018learning}\cite{guo2022tm2t}\cite{jiang2023motiongpt} learn the mutual mapping between human motion sequences and natural language descriptions. These enable describing complex motion sequences with accurate language descriptions.\par

\paragraph{Planning with LLMs}
Creating effective plans within specific environments or scenarios remains a complex task. However, with the rapid advancement of LLMs and the evolving concept of agents, planning using natural language descriptions is gaining traction. Works like \cite{huang2022language, song2023llm, liu2023llm+, sun2023adaplanner, park2023generative} have shown the potential of using LLMs as task planners across various fields. These models, even without fine-tuning, can perform diverse tasks when provided with well-crafted prompts and instructions. Nonetheless, designing these prompts and instructions requires careful thought, and often, complex pipelines are needed to achieve the desired outcomes from LLMs. While these approaches are promising for task planning, they generally do not encompass task execution, necessitating additional modules for this purpose. A recent development \cite{xiao2023unified} introduced an LLM-based framework that combines task planning with control, advancing the field of generalized human motion generation. Yet, this method can only take high-level tasks as input and produce corresponding movements, but not capable of reversing this process, limiting its utility and adaptability.\par

\section{Method}\label{sec:method}

\begin{figure*}[h]
    \centering
    \includegraphics[width=\linewidth]{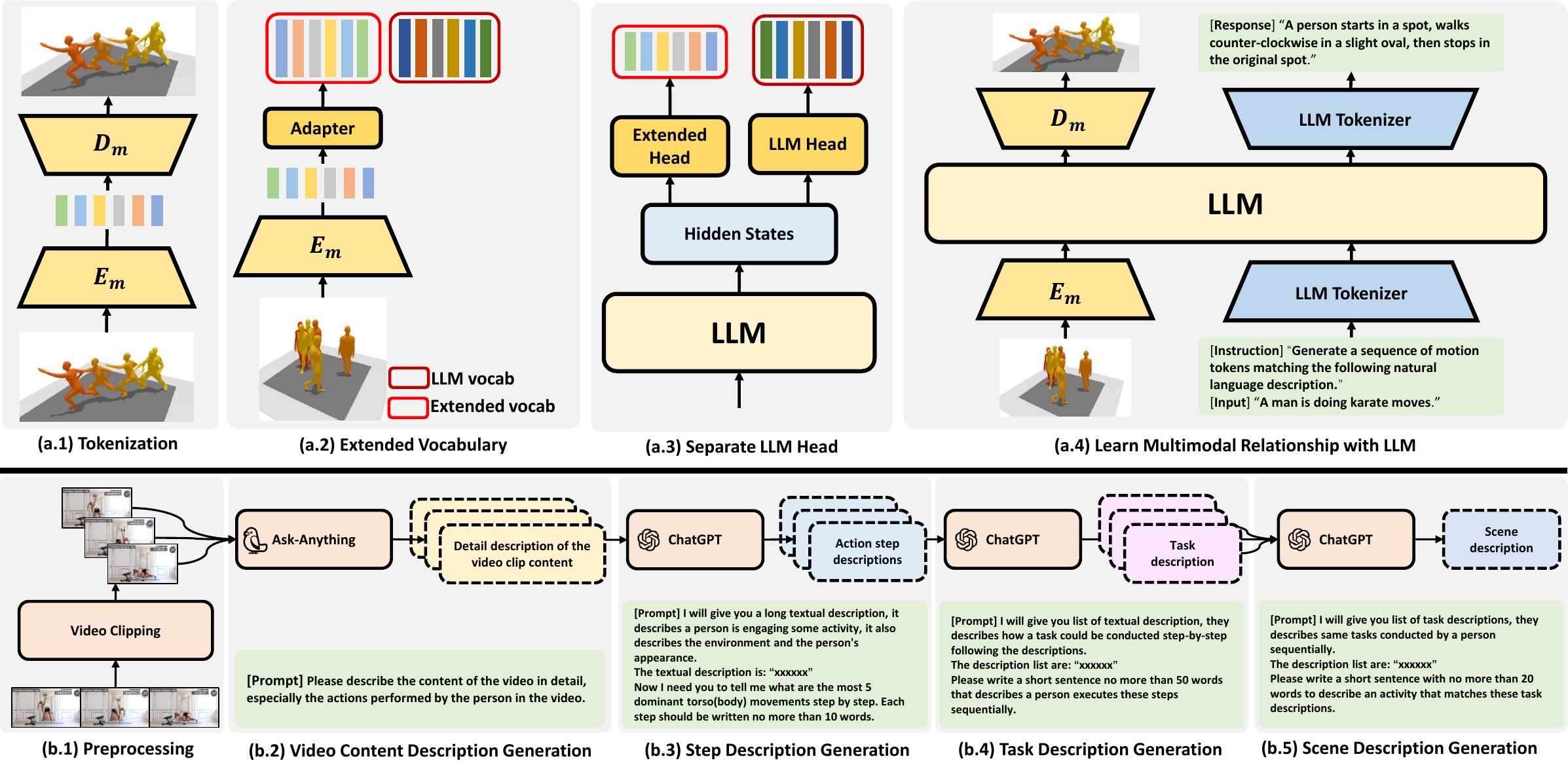}
    \caption{\textbf{Upper(a):} Multimodal LLMs as well as its detailed components. \textbf{Bottom(b):} Unsupervised Annotation Pipeline.}
    \label{fig:pipeline-method}
\end{figure*}

The overview design of our All-in-One model is shown in Fig. \ref{fig:pipeline-1}, and the technical illustration of the main components is presented in Fig. \ref{fig:pipeline-method}. Our proposed method contains two modules, 1) a Multimodal LLM that learns various relationships between text descriptions and motion sequences, and 2) a novel automatic annotation pipeline that can annotate the content of any in-the-wild videos in natural language descriptions of various levels of detail. We describe their details in the following.\par

\subsection{Motion Tokenization}We first learn a motion tokenizer to quantize continuous motion representations to discrete tokens. We follow \cite{zhang2023t2m} to train the VQ-VAE using the objective Eq. \ref{eq:vqvae}. Where $\|\Tilde{x} - x\|$ is the reconstruction term, $\|sg[z] - z_q\|$ is the embedding term, and $\|z - sg[z_q]\|$ is the commitment term.\par
\vspace{-3mm}

\begin{equation}
    \mathcal{L}_{VQ} = \|\Tilde{x} - x\| + \beta_1\|sg[z] - z_q\| + \beta_2\|z - sg[z_q]\|
    \label{eq:vqvae}
\end{equation}\par

\subsection{Motion Vocabulary} To leverage an LLM, one needs to convert continuous motion sequences into discrete tokens. One straightforward solution is using VQ-VAEs. Given a motion sequence $x \in \mathbb{R}^{T \times c}$, its tokenized representation is a set of indices $[t_0, t_1, ..., t_k]$. It is convenient to use part of the default vocabulary of LLMs as motion tokens in training LLMs\cite{zhang2023motiongpt}. Another solution is to extend the default vocabulary of LLMs and use this extended part for motion tokens\cite{jiang2023motiongpt}. However, these methods have shortcomings. For instance, \cite{zhang2023motiongpt} re-uses partial default vocabulary, which confuses the semantic meaning of associated embeddings since these embeddings are shared by multiple modalities. Although this problem is avoided in \cite{jiang2023motiongpt}, new vocabulary needs to be learned from scratch, which is inefficient especially when training on limited data. In our paper, we propose a lightweight vocabulary adaptor. It, on one hand, harnesses the semantic representation ability of discrete embeddings obtained by VQ-VAE, on the other hand, avoids learning extended vocabulary from scratch. As shown in Fig. \ref{fig:pipeline-method} (a.2), we use vq-encoder and quantizer to transform input motion sequence $x \in \mathbb{R}^{T \times c}$ to discrete embedding sequences $z^q \in \mathbb{R}^{T^q \times d}$, where $T^q$ is the length of embedding sequence, and $d$ is the embedding dimension. Because the latent space of $z^q_i \in \mathbb{R}^d$ and the hidden space $h_i \in \mathbb{R}^D$ of LLMs do not align by default, we learn an adapter layer to transform $z^q_i$ to align with $h_i$ as: 

\begin{equation}
    f_{\theta_a}(z^q): \mathbb{R}^d \rightarrow \mathbb{R}^D
    \label{eq:adapter}
\end{equation}

We show in experiments that this technique brings performance gain while reducing the model size.\par

\subsection{Separate Head for Motion Prediction}We propose to use a separate LLM head to map hidden states of LLM to motion tokens while maintaining the shape of its original head unchanged for text token prediction only. This is shown in Fig. \ref{fig:pipeline-method} (a.3). This is different from previous methods \cite{jiang2023motiongpt, zhang2023motiongpt}. In \cite{zhang2023motiongpt}, they remain the shape of the LLM head, but they re-use part of the LLM's vocabulary for motion tokens. In \cite{jiang2023motiongpt}, they modify the shape of LLM's vocabulary because of the extended vocabulary. Using a shared LLM head for tasks of different modalities brings noticeable disadvantages. Because the vocabulary is not fully shared by different modalities, it is possible to sample tokens from the vocabulary that is out of the valid range of a particular modality. Consequently, it is not guaranteed that any sampled token can be decoded as expected. Our method solves this problem. We denote the original vocabulary of LLM as $\mathcal{V}_t=\{v_i\}_{i=1}^N$, and the extended vocabulary for motion is $\mathcal{V}_m=\{v_0\}_{i=1}^M$, where $M$ and $N$ are vocabulary sizes and $M \neq N$. Given hidden states of LLM $h_i \in \mathbb{R}^D$, the original LLM head is $f_{\theta_t}(h): \mathbb{R}^D \rightarrow \mathbb{R}^N$, and the separate head for motion is $f_{\theta_m}(h): \mathbb{R}^D \rightarrow \mathbb{R}^M$. Therefore, learning modality-specific heads guarantees we can always sample tokens from the correct vocabulary. Hence, decoding performs expectedly.\par

\subsection{Instruction Tuning the LLMs}Fig.~\ref{fig:pipeline-method} (a.4) shows that an LLM could be used to learn motion-related multimodal tasks tuned by different instructions. The fundamental concept of LLMs is to treat all inputs as discrete tokens and predict subsequent tokens based on previous ones as $p_{\theta}(x_i | x_{<i})$. In our case, there are two modalities, text and motion. We use LLM's original vocabulary for text modality. For motion sequences, we first transform them to discrete embeddings as $z_q=\mathcal{Q}(\mathcal{E}(x))$, where $\mathcal{E}(\cdot)$ is the vq-encoder, and $\mathcal{Q}(\cdot)$ is the quantizer. Then we map the motion embeddings to extended vocabulary embeddings using Eq. \ref{eq:adapter}. Through this process, we can align two different modalities into a cohesive and LLM-favorable representation, allowing us to harness the ability of LLMs to learn complex relationships between these modalities. Hence, we can formulate motion-related multimodal tasks as conditioned language generation problems. 

Let's denote the condition as a sequence of tokens as $C=\{c_i\}_{i=1}^{K_c}$, and the target as a sequence of tokens as $X=\{x_i\}_{i=1}^{K_x}$, the condition and target could be either text or human motion, and we can model the conditioned generation problem as $p_{\theta}(x_i|x_{i<i}, C)$. We use a transformer encoder to extract the context information from condition $C$. At this stage, the full attention mechanism is adopted because we found it is more favorable than using causal attention in learning contextual information from the condition. Then we use a transformer decoder with causal attention to learn the relationship between target tokens and conditions. We found T5's\cite{raffel2020exploring} encoder-decoder architecture is an appropriate choice. Because our model predicts the probability distribution of target tokens at every step, we use cross-entropy loss to supervise the fine-tuning. Since we use separate heads for text and motion modalities, our objectives for both modalities are as follows:

\begin{equation}
    \mathcal{L}_t = -\Sigma_{i=1}^T{\hat{x}_i\log(p_{\theta,\theta_t}(x_i|x_{<i},C))}
    \label{eq:ce_text}
\end{equation}

\begin{equation}
    \mathcal{L}_m = -\Sigma_{i=1}^T{\hat{x}_i\log(p_{\theta,\theta_m}(x_i|x_{<i},C))}
    \label{eq:ce_motion}
\end{equation}

Where $\theta$, $\theta_t$, and $\theta_m$ are parameters of LLM, original LLM head, and extended LLM head, respectively. We list the prompts for instruction tuning in Appendix A.

\begin{figure*}
    \centering
    \includegraphics[width=1.0\linewidth]{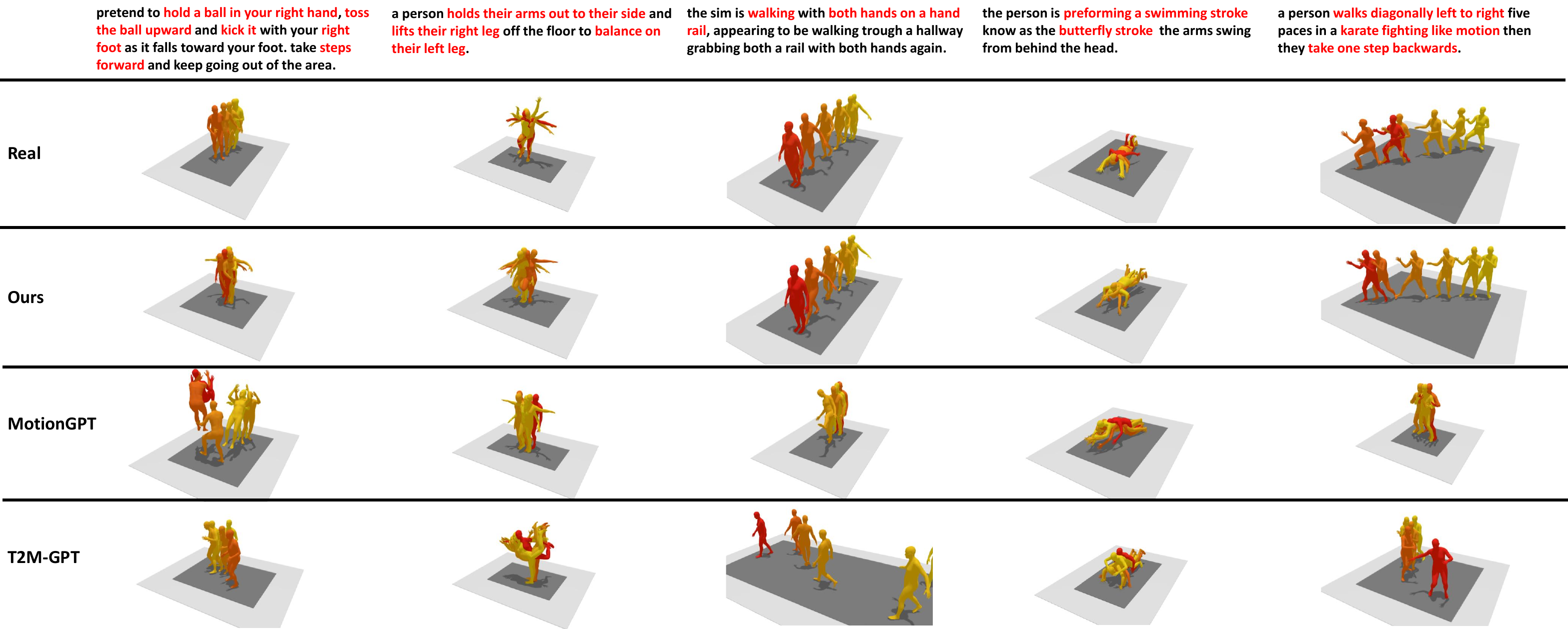}
    \caption{\textbf{Comparison of Motion Generation.} We compare the motion generation between ours and SOTA\cite{jiang2023motiongpt, zhang2023t2m}. We highlight the keywords to ease the visualization.}
    \label{fig:low-t2m}
\end{figure*}

\begin{figure*}
    \centering
    \includegraphics[width=1.0\linewidth]{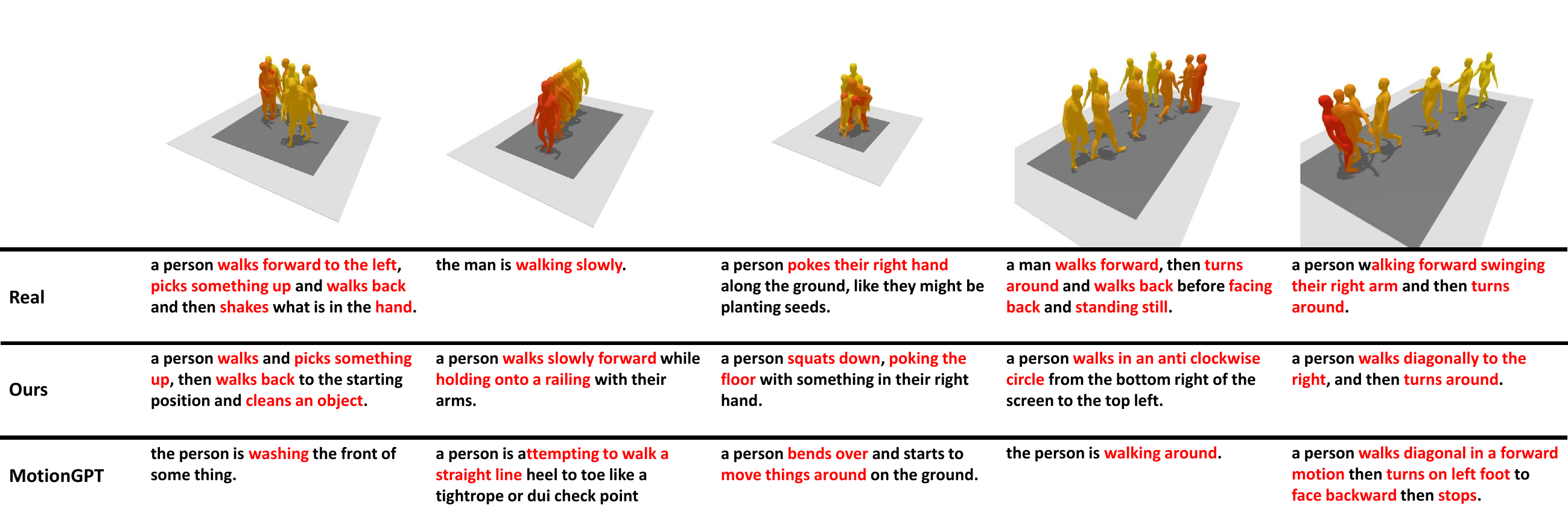}
    \caption{\textbf{Comparison of Motion Understanding.} We compare the motion understanding performance between ours and MotionGPT\cite{jiang2023motiongpt}. We highlight the keywords to show the alignment between motion and text.}
    \label{fig:low-m2t}
\end{figure*}

\begin{table*}
    \centering
    \resizebox{1.0\linewidth}{!}{
        \begin{tabular}{c|cccccc|ccccc|ccc}
            \Xhline{2pt}
            \multirow{2}{*}{Method} & \multicolumn{6}{c|}{Motion Generation} & \multicolumn{5}{c|}{Motion Understanding} & \multicolumn{3}{c}{Motion-in-Between} \\
            \cline{2-15}
             & R Top-1 $\uparrow$ & R Top-2 $\uparrow$ & R Top-3 $\uparrow$ & FID $\downarrow$ & Div $\rightarrow$ & MM $\uparrow$ & Blue-1 $\uparrow$ & Blue-4 $\uparrow$ & Rouge $\uparrow$ & Cider $\uparrow$ & BertScore $\uparrow$ & FID $\downarrow$ & Div $\rightarrow$ & MM $\uparrow$ \\
            \hline
            GT & 0.511 & 0.703 & 0.797 & $0.002^{\pm{0.000}}$ & $9.503^{\pm{0.065}}$ & - & - & - & - & - & - & 0.002 & 9.503 & - \\
            \hline
            TM2T\cite{guo2022tm2t} & $0.424^{\pm{0.003}}$ & $0.618^{\pm{0.003}}$ & $0.729^{\pm{0.002}}$ & $1.501^{\pm{0.017}}$ & $8.589^{\pm{0.076}}$ & $2.424^{\pm{0.093}}$ & \cellcolor{yellow!15}48.90 & 7.00 & \cellcolor{yellow!15}38.10 & 16.80 & 32.20 & - & - & - \\
            MDM\cite{tevet2022human} & $0.320^{\pm{0.005}}$ & $0.498^{\pm{0.004}}$ & $0.611^{\pm{0.007}}$ & $0.544^{\pm{0.044}}$ & \cellcolor{yellow!15}$9.559^{\pm{0.086}}$ & $2.799^{\pm{0.072}}$ & - & - & - & - & - & 2.698 & 8.42 & - \\
            MLD\cite{chen2023executing} & $0.481^{\pm{0.003}}$ & $0.673^{\pm{0.003}}$ & $0.772^{\pm{0.002}}$ & $0.473^{\pm{0.013}}$ & $9.724^{\pm{0.082}}$ & $2.413^{\pm{0.079}}$ & - & - & - & - & - & - & - & - \\
            T2M-GPT\cite{zhang2023t2m} & $0.491^{\pm{0.005}}$ & $0.680^{\pm{0.003}}$ & $0.775^{\pm{0.002}}$ & \cellcolor{red!15}$0.116^{\pm{0.004}}$ & $9.761^{\pm{0.081}}$ & $1.856^{\pm{0.011}}$ & - & - & - & - & - & - & - & - \\
            MotionGPT\cite{zhang2023motiongpt} & $0.411^{\pm{0.000}}$ & $0.594^{\pm{0.000}}$ & $0.696^{\pm{0.000}}$ & $0.542^{\pm{0.000}}$ & $9.311^{\pm{0.000}}$ & - & - & - & - & - & - & - & - & - \\
            MotionGPT\cite{jiang2023motiongpt} & \cellcolor{yellow!15}$0.492^{\pm{0.003}}$ & \cellcolor{yellow!15}$0.681^{\pm{0.003}}$ & \cellcolor{yellow!15}$0.778^{\pm{0.002}}$ & $0.232^{\pm{0.008}}$ & \cellcolor{red!15}$9.528^{\pm{0.071}}$ & $3.096^{\pm{0.008}}$ & 48.20 & \cellcolor{yellow!15}12.47 & 37.40 & \cellcolor{yellow!15}29.20 & \cellcolor{yellow!15}32.40 & \cellcolor{red!15}0.214 & \cellcolor{red!15}9.56 & - \\
            HMD\cite{shafir2023human} & - & - & - & - & - & - & - & - & - & - & - & \cellcolor{yellow!15}1.48 & 8.90 & - \\
            \hline
            Ours & \cellcolor{red!15}$0.510^{\pm{0.005}}$ & \cellcolor{red!15}$0.702^{\pm{0.005}}$ & \cellcolor{red!15}$0.796^{\pm{0.003}}$ & \cellcolor{yellow!15}$0.168^{\pm{0.0083}}$ & $9.624^{\pm{0.0545}}$ & - & \cellcolor{red!15}49.28 & \cellcolor{red!15}12.70 & \cellcolor{red!15}40.44 & \cellcolor{red!15}32.65 & \cellcolor{red!15}53.58 & $1.655^{\pm{0.020}}$ & \cellcolor{yellow!15}$9.015^{\pm{0.095}}$ & \cellcolor{red!15}$7.417^{\pm{0.662}}$ \\
            \Xhline{2pt}
        \end{tabular}
    }
    \caption{\textbf{Results of Low-level Tasks.} We compare our method with various SOTAs on low-level tasks such as 1) Motion Generation, 2) Motion Understanding, and 3) Motion-in-Between. $\colorbox{red!15}{\rm Indicate best results}, \colorbox{yellow!15}{\rm indicates second best results}$.}
    \label{tab:main-low-level}
\end{table*}

\begin{table*}[]
    \centering
    \resizebox{0.6\linewidth}{!}{
        \begin{tabular}{c|cccccccc}
            \Xhline{2pt}
            Method & CT2T $\uparrow$ & CS2S $\uparrow$ & CT2S $\uparrow$ & CS2T $\uparrow$ & T2C $\uparrow$ & S2C $\uparrow$ & T2S $\uparrow$ & S2T $\uparrow$  \\
            \hline
            GT & 0.843 & 0.882 & 0.818 & 0.849 & 0.872 & 0.910 & 0.678 & 0.812 \\
            \hline
            Ours(GPT2-Large) &  0.751 & 0.786 & 0.705 & 0.784 & 0.873 & \cellcolor{red!15}0.895 & 0.956 & 0.955 \\
            Ours(T5-Large) & \cellcolor{red!15}0.843 & \cellcolor{red!15}0.937 & \cellcolor{red!15}0.884 & \cellcolor{red!15}0.841 & \cellcolor{red!15}0.940 & 0.893 & \cellcolor{red!15}0.994 & \cellcolor{red!15}0.997 \\
            \Xhline{2pt}
        \end{tabular}
    }
    \caption{\textbf{Results of High-level Tasks.} We evaluate the Logical Coherent Score(LCS) on 8 high-level tasks, and we compare the results of our method by using T5 and GPT architecture. }
    \label{tab:main-high-level}
\end{table*}

\begin{figure*}
    \centering
    \includegraphics[width=1.0\linewidth]{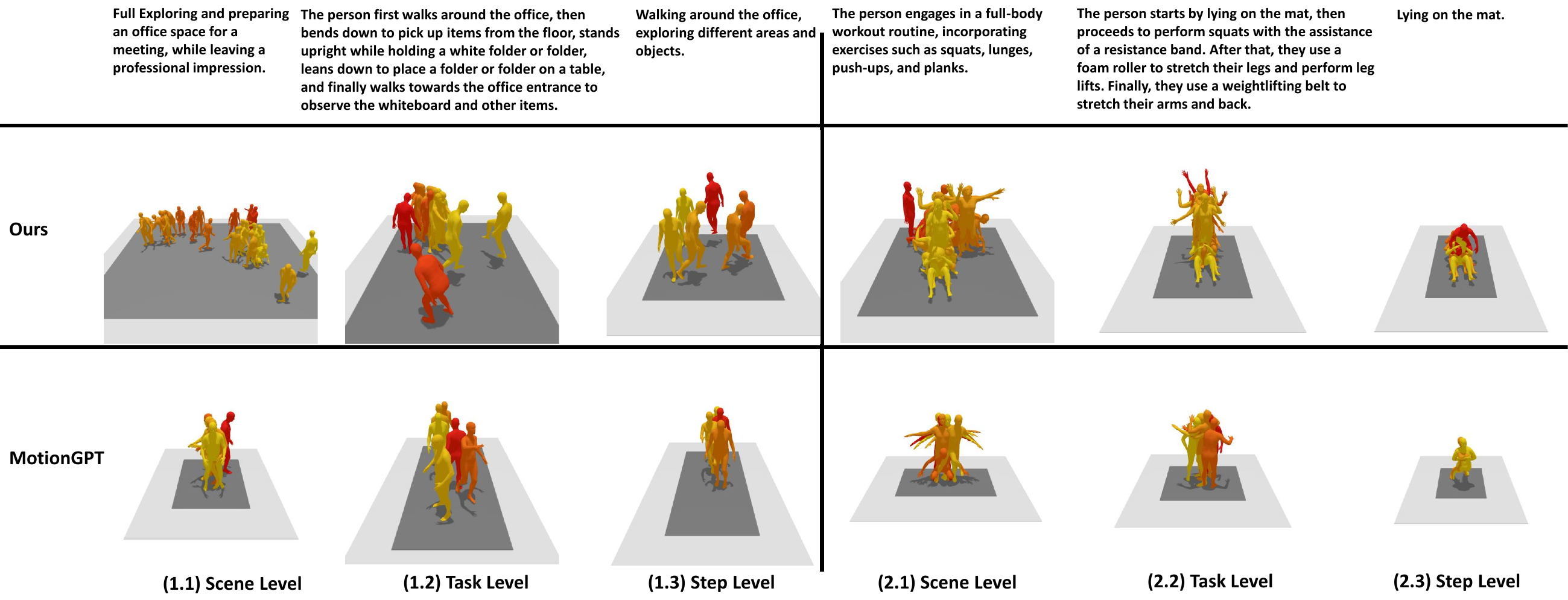}
    \caption{\textbf{Comparison of Motion Generation with Various Text Granularity.} We compare the generation results conditioned on coarse-grained(scene), middle-grained(task), and fine-grained(step) descriptions. Our method generates long motion with 2K+ frames from scene description(1.1, 2.1) and 0.6K frames from task description(1.2, 2.2), while MotionGPT\cite{jiang2023motiongpt} only results 0.2K+ frames for both scenarios.}
    \label{fig:full-t2m}
\end{figure*}

\subsection{Automatic Annotation Pipeline}\label{sec:3.5}We present a novel yet efficient method to annotate textual descriptions from in-the-wild videos of various levels of detail. Our method is shown in Fig. \ref{fig:pipeline-method} (b). We use this pipeline to collect datasets to fine-tune our model on tasks such as task planning, decomposition, scene estimation, etc. Given any video, we first crop it into segments of fixed length, then we use a Visual-LLM to describe its content as detail as possible. In our pipeline, we adopt \cite{2023videochat} for its encouraging ability in multimodal understanding and text generation. We then leverage ChatGPT\cite{chatgpt} to describe how the person in the video moves step by step based on the very detailed descriptions, and we define descriptions of this level of detail: \textbf{step description}. This is the most fine-grained description, and each video segment could contain multiple steps. We also use ChatGPT to extract one summarized description for each video segment from the descriptions of the steps as well, which is denoted as: \textbf{task description}. In addition, we define the description that depicts the scene information of the entire video as: \textbf{scene description}, which is the most coarse-grained among all. To obtain this description, we feed the task descriptions related to each video segment to ChatGPT sequentially, and again ask ChatGPT to summarize the scene description that matches these tasks coherently. We describe the prompt structures used in this pipeline in Appendix B.\par

\section{Experiments}\label{sec:experiments}

\begin{figure*}
    \centering
    \includegraphics[width=1.0\linewidth]{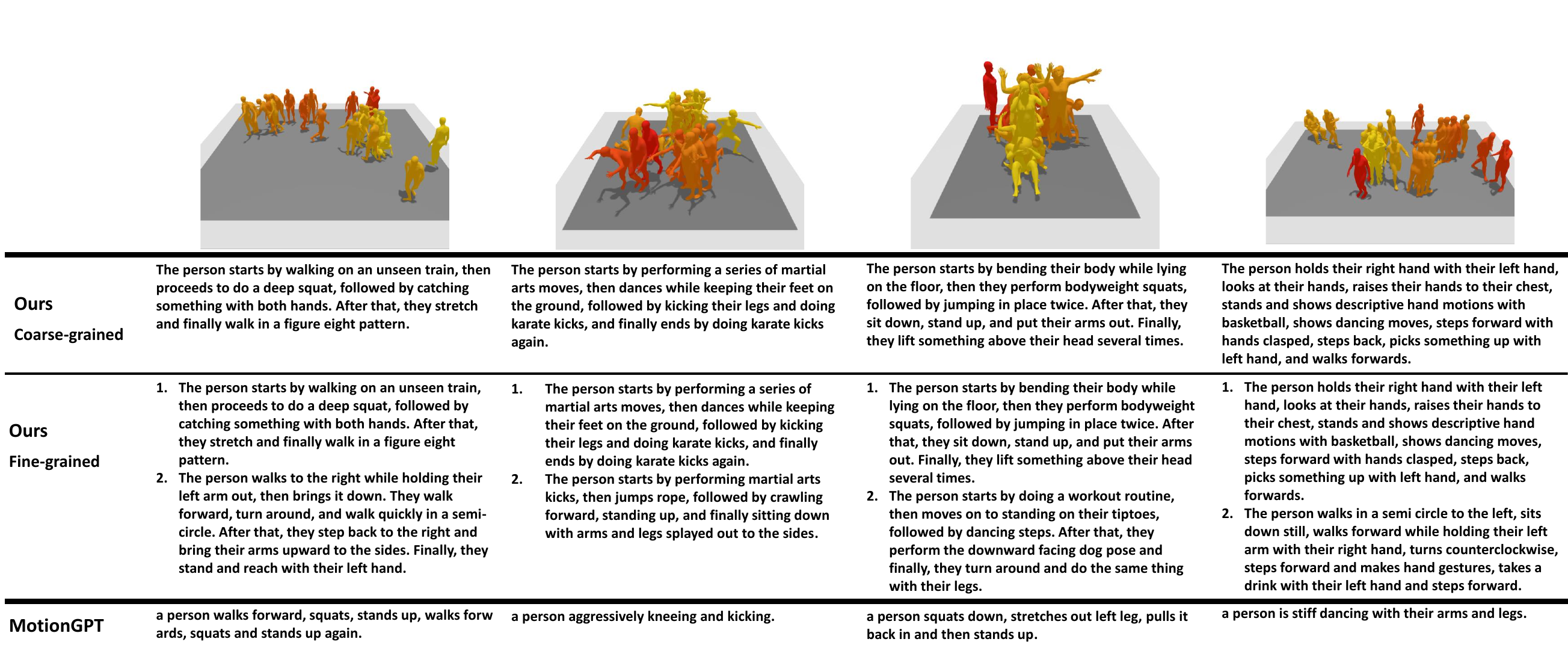}
    \caption{\textbf{Comparison of Motion Understanding of Various Level of Detail.} We compare the long motion understanding in various levels of detail with MotionGPT\cite{jiang2023motiongpt}. The motions have around 2K+ frames in length. Our method is able to describe the motion at both coarse- and fine-grained levels of detail.}
    \label{fig:full-m2t}
\end{figure*}

\subsection{Datasets}We use HumanML3D\cite{guo2022tm2t} to train and evaluate our method on three low-level tasks, including motion generation, motion understanding, and motion-in-between synthesis. The data preprocessing follows \cite{guo2022tm2t}. Because there are no appropriate datasets for motion task planning, scene understanding, task decomposition, etc, we collect a new dataset using the pipeline proposed in Sec. \ref{sec:3.5}. The detailed procedure of dataset collection, the format, and the structure of the dataset are described in Appendix B, F.

\subsection{Evaluation Metrics}

\begin{table*}
    \centering
    \resizebox{0.8\linewidth}{!}{
        \begin{tabular}{c|ccccc|ccccc}
            \Xhline{2pt}
            \multirow{2}{*}{Method} & \multicolumn{5}{c|}{Motion Understanding} & \multicolumn{5}{c}{Task Sumarization} \\
            \cline{2-11}
             & Blue-1 $\uparrow$ & Blue-4 $\uparrow$ & Rouge $\uparrow$ & Cider $\uparrow$ & BertScore $\uparrow$ & Blue-4 $\uparrow$ & Rouge $\uparrow$ & Cider $\uparrow$ & BertScore $\uparrow$ \\
             \hline
             Ours(GPT2-Large) & \cellcolor{red!15}33.83 & \cellcolor{red!15}1.39 & \cellcolor{red!15}20.71 & \cellcolor{red!15}1.49 & 39.41 & 34.82 & 2.19 & 16.76 & 2.11 & 26.39 \\
             Ours(T5-Large) & 30.14 & 1.22 & 17.85 & 1.40 & \cellcolor{red!15}41.05 & \cellcolor{red!15}40.11 & \cellcolor{red!15}7.45 & \cellcolor{red!15}24.82 & \cellcolor{red!15}3.03 & \cellcolor{red!15}32.28 \\
            \Xhline{2pt}
        \end{tabular}
    }
    \caption{\textbf{Results of Full Pipeline Planning and Generation.} We assess our method's performance on a full pipeline based on the concept of cycle consistency. We adopt linguistic similarity metrics to evaluate the task-level and step-level consistency. }
    \label{tab:main-cycle}
\end{table*}

\begin{table}[]
    \centering
    \resizebox{1.0\linewidth}{!}{
        \begin{tabular}{c|cc|cc|c}
            \Xhline{2pt}
            \multirow{2}{*}{Method} & \multicolumn{2}{c|}{LCS} & \multicolumn{2}{c|}{Ling. Consis.} & \multirow{2}{*}{T2M Consis $\uparrow$.} \\
            \cline{2-5}
             & CT2T $\uparrow$ & T2S $\uparrow$ & Step $\uparrow$ & Task $\uparrow$ & \\
            \hline
            Ours(GPT2-Large) & 88.50 & 92.32 & 55.56 & 65.00 & 2.8 \\
            Ours(T5-Large) & \cellcolor{red!15}90.67 & \cellcolor{red!15}97.87 & \cellcolor{red!15}56.68 & \cellcolor{red!15}92.86 & \cellcolor{red!15}4.02 \\
            \Xhline{2pt}
        \end{tabular}
    }
    \caption{\textbf{Results of User Study.} We assess our method's performance in terms of task planning, decomposition, motion generation, understanding, and task summarization. `LCS', `Ling. Consis.', and `T2M Consis.' respectively denote logical coherent score, linguistic consistency, and text-to-motion consistency.}
    \label{tab:user-study}
\end{table}

\begin{table*}
    \centering
    \resizebox{1.0\linewidth}{!}{
        \begin{tabular}{c|ccccc|ccccc|cc}
            \Xhline{2pt}
            \multirow{2}{*}{Method} & \multicolumn{5}{c|}{Motion Generation} & \multicolumn{5}{c|}{Motion Understanding} & \multicolumn{2}{c}{Motion-in-Between} \\
            \cline{2-13}
             & R Top-1 $\uparrow$ & R Top-1 $\uparrow$ & R Top-3 $\uparrow$ & FID $\downarrow$ & Div $\rightarrow$ & Bleu-1 $\uparrow$ & Bleu-4 $\uparrow$ & Rouge $\uparrow$ & Cider $\uparrow$ & BertScore $\uparrow$ & FID $\downarrow$ & Div $\rightarrow$ \\
            \hline
            Llama-13B & 0.389 & 0.539 & 0.623 & 0.567 & \cellcolor{red!15}9.489 & 49.01 & 12.42 & 39.01 & 27.73 & 51.31 & - & - \\
            GTP2-Large & 0.454 & 0.635 & 0.728 & 0.316 & 10.021 & \cellcolor{red!15}51.92 & \cellcolor{red!15}13.29 & \cellcolor{red!15}40.93 & \cellcolor{red!15}33.69 & 51.73 & \cellcolor{red!15}0.911 & 9.007 \\
            T5-Base & 0.468 & 0.654 & 0.751 & 0.284 & 9.967 & 48.29 & 11.51 & 38.51 & 28.11 & 51.59 & 4.042 & 8.982 \\
            T5-Large & \cellcolor{red!15}0.510 & \cellcolor{red!15}0.702 & \cellcolor{red!15}0.796 & \cellcolor{red!15}0.168 & 9.624 & 49.28 & 12.70 & 40.44 & 32.65 & \cellcolor{red!15}53.58 & 1.655 & \cellcolor{red!15}9.015 \\
            \hline
            w/o adapter & 0.497 & 0.688 & 0.784 & 0.215 & 9.842 & 50.42 & 11.75 & 37.29 & 27.23 & 49.34 & 1.074 & 8.818 \\
            w/ adapter & \cellcolor{red!15}0.510 & \cellcolor{red!15}0.702 & \cellcolor{red!15}0.796 & \cellcolor{red!15}0.168 & \cellcolor{red!15}9.624 & \cellcolor{red!15}49.28 & \cellcolor{red!15}12.70 & \cellcolor{red!15}40.44 & \cellcolor{red!15}32.65 & \cellcolor{red!15}53.58 & \cellcolor{red!15}1.655 & \cellcolor{red!15}9.015 \\
            \Xhline{2pt}
        \end{tabular}
    }
    \caption{\textbf{Ablation Study.} We investigate the effectiveness of model architecture, sizes, and ways of introducing extended vocabulary. }
    \label{tab:ablation}
\end{table*}

\paragraph{Metrics for Low-level Tasks}We use R-Precision score\cite{guo2022tm2t} to measure the alignment between textual description and generated motion sequence, Frechet Inception Distance(FID) and Diversity(Div) to measure the generation quality,  Multimodality(MM) to evaluate the diversity of generation driven by the same control signal. FID, Div, and MM are used for motion generation and motion-in-between tasks as well. To assess the motion understanding performance, we measure the linguistic similarity between annotated texts and generated texts using the following metrics: 1) BLUE\cite{papineni2002bleu}, 2) BLUE\cite{papineni2002bleu}, 3) ROUGE-L\cite{lin2004rouge}, 4) CIDEr\cite{vedantam2015cider}. Refer to Appendix C for detailed descriptions.

\paragraph{Metrics for High-level Tasks}We propose a novel metric, Logical Coherence Score(LCS), to evaluate the performance of high-level tasks, such as task planning, decomposition, and scene estimation, because there are no proper metrics for automatic assessment. Unlike motion understanding, where we can measure the semantic similarity between candidate and reference, there is no single correct answer for these high-level tasks. Whether the planned task is logically coherent with historical tasks or scenes is the primary concern. We therefore leverage the knowledge and reasoning ability of ChatGPT to judge whether our generated tasks match the conditions logically. For every generated text, we ask ChatGPT to determine whether it is logically coherent with conditions or not as $\mathbbm{1}(\hat{x}_i, x_i) = 1$ if $\hat{x}_i$ and $x_i$ are logically coherent, otherwise  $\mathbbm{1}(\hat{x}_i, x_i) = 0$. Therefore, the coherent score for every task is calculated as $LCS = \frac{1}{N}\Sigma_{i=1}^N(\mathbbm{1}(\hat{x}_i, x_i))$. We present the workflow of evaluation, as well as the specific prompts for each task in Appendix D.\par

\paragraph{Metrics for Full Pipeline}We define the full pipeline as: given a scene description, it conducts multiple rounds of task planning, decomposes each task to up to 5 steps descriptions, then synthesizes motions out of these steps, and finally blends these motions to one long sequence. Since there is no ground truth data, it is difficult to evaluate directly, we propose to evaluate the full pipeline based on the concept of cycle consistency. Concretely, we generate a long motion sequence from multiple tasks $x_t^f$ and corresponding step descriptions $x_s^f$. Then we conduct a motion understanding task to describe the motion in various text granularity, resulting task- and step-level descriptions as $x_t^r$, $x_s^r$ respectively. Ideally, $x_t^f$ and $x_t^r$ should describe the same action, and so do $x_s^f$ and $x_s^r$. We use BertScore, BLUE, ROUGE-L, and CIDEr to evaluate their linguistic similarity. We also conducted a user study to measure logical coherence, linguistic consistency, and motion-to-text consistency. Please refer to Appendix E for more details.

\subsection{Results}

\paragraph{Results of Low-level Tasks} HumanML3D\cite{guo2022tm2t} is used to jointly train three low-level tasks. For testing,  we generate 10 samples from every condition signal and report their mean and 95\% confidence interval. We compare our method with various SOTAs and present the quantitative results in Tab. \ref{tab:main-low-level}. The comparison shows that our method largely outperforms all previous approaches on motion understanding tasks. Our method also ranks first in terms of R-Precision metrics on the motion generation task and achieves competitive results on other metrics. We also compare our method on the motion-in-between task with different SOTAs and observe competitive performance.  
\vspace{-3mm}

\paragraph{Results of High-level Tasks}We define 8 subtasks as high-level tasks. Denote `C' as scene(context) description, `T' as task description, and `S' as step description, task `CT2T' stands for given scene(T) and one historical task(T) description, conduct one round of planning to predict the next possible task(T). Following this notation, we define the following 8 subtasks as CT2T, CS2S, CT2S, CS2T, T2C, S2C, T2S, and S2T. Because there are no similar methods designed for these tasks, we present a novel benchmark(Tab. \ref{tab:main-high-level}) in this paper for two purposes: 1) we measure the Logical Coherent Score(LCS) on our collected dataset and the generation results of our method to justify the effectiveness of our method, and 2) we provide this benchmark and dataset for future research. We use T5-Large and GPT2-Large as our LLM and report the quantitative metrics on these 8 subtasks respectively, and the results suggest that using T5-Large outperforms GPT2-Large on 7 of 8 tasks.
\vspace{-3mm}

\paragraph{Results of Full Pipeline}Our model supports 1) generating very long motion sequences from a single scenario description and 2) describing the long motion at various text granularities. The depiction of motion should match the planned task/step description semantically. Therefore, we evaluate the consistency using \cite{zhang2019bertscore, papineni2002bleu, vedantam2015cider, lin2004rouge} on task summarization(task-level) and motion understanding(step-level) cycle consistency. We compare two variants of LLM architecture and report the results in Tab. \ref{tab:main-cycle}. The results suggests that using T5-Large as LLM outperforms GPT2-Large significantly on task summarization, and achieves competitive results on motion understanding tasks.
\vspace{-3mm}

\paragraph{User Study}We conduct user study to investigate the performance of our method. Given a scene description, we run task planning, decomposition, motion synthesis, and motion-in-between, resulting in descriptions in various detail granularities and a long motion sequence. We then perform motion understanding and task summarization to describe the motion at fine- and coarse-grained. We evaluate the 1) LCS on planning and decomposition, and 2) linguistic consistency on motion understanding and summarization tasks. Participants are expected to score '1' to correct(coherent and consistent) answers, while '0' to the incorrect. In addition, human evaluation on the consistency between generated long motion and descriptions is also performed, with a rating scale of 1-5 as a metric. The results are reported in Tab. \ref{tab:user-study}. The results suggest using T5-Large as LLM has stronger ability in planning, decomposition, synthesis, understanding, summarization, and in-between tasks than GPT2-Large. The conclusions of manual and automated evaluation demonstrate high degree of consistency, supporting the effectiveness of our method.

\subsection{Ablation Study}We conduct various ablation studies to investigate the effectiveness of 1) LLM model architecture, 2) LLM model size, and 3) extended vocabulary with an adapter.
\vspace{-3mm}

\paragraph{LLM Model Architecture}We investigate the effectiveness of various LLM model architectures. We adopt Llama-13B\cite{touvron2023llama}, GPT2-Large\cite{radford2019language}, and T5-Large\cite{raffel2020exploring} and conduct multitask fine-tuning. LoRA\cite{hu2022lora} is adopted for Llama-13B to save memory usage. The results on low-level tasks are shown in Tab. \ref{tab:ablation}, and those of high-level tasks are reported in Tab. \ref{tab:main-high-level} as well. We found that Llama-13B has poor performance on generation and understanding, this is because LoRA actually finetunes $<1\%$ parameters to connect motion modality with text, which fails to fully use the capacity of Llama. Although GTP2-Large has slightly better performance on motion understanding, T5-Large outperforms on all other tasks, suggesting the superiority of T5-Large against GPT2-Large as LLM architecture for our scenario.
\vspace{-3mm}

\paragraph{LLM Model Sizes}We also compare the performance discrepancy in terms of model sizes. Tab. \ref{tab:ablation} shows that a larger model size(T5-Large) brings noticeable performance gain.
\vspace{-3mm}

\paragraph{Extended Vocab. w/ or w/o Adapter}We investigate the effectiveness of different approaches to introducing extended vocabulary. 1) Extend the LLM vocabulary and learn the weights from scratch. 2) Align the quantized embeddings to the LLM vocabulary embedding space through the adapter layer. Tab. \ref{tab:ablation} demonstrates that our proposed approach achieves remarkable improvements on various tasks.

\section{Conclusion}
We show the synergy of unifying seven motion-related tasks through our newly introduced All-in-One framework. For the first time, we demonstrate that our method can enable long-motion synthesis thanks to its iterative planning, understanding, generation, etc, within the integral framework.   

{
    \small
    \bibliographystyle{ieeenat_fullname}
    \bibliography{main}

\begin{thebibliography}{44}
\providecommand{\natexlab}[1]{#1}
\providecommand{\url}[1]{\texttt{#1}}
\expandafter\ifx\csname urlstyle\endcsname\relax
  \providecommand{\doi}[1]{doi: #1}\else
  \providecommand{\doi}{doi: \begingroup \urlstyle{rm}\Url}\fi

\bibitem[Brown et~al.(2020)Brown, Mann, Ryder, Subbiah, Kaplan, Dhariwal, Neelakantan, Shyam, Sastry, Askell, et~al.]{brown2020language}
Tom Brown, Benjamin Mann, Nick Ryder, Melanie Subbiah, Jared~D Kaplan, Prafulla Dhariwal, Arvind Neelakantan, Pranav Shyam, Girish Sastry, Amanda Askell, et~al.
\newblock Language models are few-shot learners.
\newblock \emph{Advances in neural information processing systems}, 33:\penalty0 1877--1901, 2020.

\bibitem[Chen et~al.(2023)Chen, Jiang, Liu, Huang, Fu, Chen, and Yu]{chen2023executing}
Xin Chen, Biao Jiang, Wen Liu, Zilong Huang, Bin Fu, Tao Chen, and Gang Yu.
\newblock Executing your commands via motion diffusion in latent space.
\newblock In \emph{Proceedings of the IEEE/CVF Conference on Computer Vision and Pattern Recognition}, pages 18000--18010, 2023.

\bibitem[Chi et~al.(2022)Chi, Ha, Chi, Lee, Huang, and Ramani]{chi2022infogcn}
Hyung-gun Chi, Myoung~Hoon Ha, Seunggeun Chi, Sang~Wan Lee, Qixing Huang, and Karthik Ramani.
\newblock Infogcn: Representation learning for human skeleton-based action recognition.
\newblock In \emph{Proceedings of the IEEE/CVF Conference on Computer Vision and Pattern Recognition}, pages 20186--20196, 2022.

\bibitem[Chung et~al.(2022{\natexlab{a}})Chung, Hou, Longpre, Zoph, Tay, Fedus, Li, Wang, Dehghani, Brahma, et~al.]{chung2022scaling}
Hyung~Won Chung, Le Hou, Shayne Longpre, Barret Zoph, Yi Tay, William Fedus, Eric Li, Xuezhi Wang, Mostafa Dehghani, Siddhartha Brahma, et~al.
\newblock Scaling instruction-finetuned language models.
\newblock \emph{arXiv preprint arXiv:2210.11416}, 2022{\natexlab{a}}.

\bibitem[Chung et~al.(2022{\natexlab{b}})Chung, Hou, Longpre, Zoph, Tay, Fedus, Li, Wang, Dehghani, Brahma, Webson, Gu, Dai, Suzgun, Chen, Chowdhery, Castro-Ros, Pellat, Robinson, Valter, Narang, Mishra, Yu, Zhao, Huang, Dai, Yu, Petrov, Chi, Dean, Devlin, Roberts, Zhou, Le, and Wei]{flann2}
Hyung~Won Chung, Le Hou, Shayne Longpre, Barret Zoph, Yi Tay, William Fedus, Yunxuan Li, Xuezhi Wang, Mostafa Dehghani, Siddhartha Brahma, Albert Webson, Shixiang~Shane Gu, Zhuyun Dai, Mirac Suzgun, Xinyun Chen, Aakanksha Chowdhery, Alex Castro-Ros, Marie Pellat, Kevin Robinson, Dasha Valter, Sharan Narang, Gaurav Mishra, Adams Yu, Vincent Zhao, Yanping Huang, Andrew Dai, Hongkun Yu, Slav Petrov, Ed~H. Chi, Jeff Dean, Jacob Devlin, Adam Roberts, Denny Zhou, Quoc~V. Le, and Jason Wei.
\newblock Scaling instruction-finetuned language models, 2022{\natexlab{b}}.

\bibitem[Guo et~al.(2020)Guo, Zuo, Wang, Zou, Sun, Deng, Gong, and Cheng]{guo2020action2motion}
Chuan Guo, Xinxin Zuo, Sen Wang, Shihao Zou, Qingyao Sun, Annan Deng, Minglun Gong, and Li Cheng.
\newblock Action2motion: Conditioned generation of 3d human motions.
\newblock In \emph{Proceedings of the 28th ACM International Conference on Multimedia}, pages 2021--2029, 2020.

\bibitem[Guo et~al.(2022)Guo, Xuo, Wang, and Cheng]{guo2022tm2t}
Chuan Guo, Xinxin Xuo, Sen Wang, and Li Cheng.
\newblock Tm2t: Stochastic and tokenized modeling for the reciprocal generation of 3d human motions and texts.
\newblock \emph{arXiv preprint arXiv:2207.01696}, 2022.

\bibitem[Hu et~al.(2022)Hu, Shen, Wallis, Allen-Zhu, Li, Wang, Wang, and Chen]{hu2022lora}
Edward~J Hu, Yelong Shen, Phillip Wallis, Zeyuan Allen-Zhu, Yuanzhi Li, Shean Wang, Lu Wang, and Weizhu Chen.
\newblock Lo{RA}: Low-rank adaptation of large language models.
\newblock In \emph{International Conference on Learning Representations}, 2022.

\bibitem[Huang et~al.(2022)Huang, Abbeel, Pathak, and Mordatch]{huang2022language}
Wenlong Huang, Pieter Abbeel, Deepak Pathak, and Igor Mordatch.
\newblock Language models as zero-shot planners: Extracting actionable knowledge for embodied agents.
\newblock In \emph{International Conference on Machine Learning}, pages 9118--9147. PMLR, 2022.

\bibitem[Jiang et~al.(2023)Jiang, Chen, Liu, Yu, Yu, and Chen]{jiang2023motiongpt}
Biao Jiang, Xin Chen, Wen Liu, Jingyi Yu, Gang Yu, and Tao Chen.
\newblock Motiongpt: Human motion as a foreign language.
\newblock \emph{arXiv preprint arXiv:2306.14795}, 2023.

\bibitem[Karunratanakul et~al.(2023)Karunratanakul, Preechakul, Suwajanakorn, and Tang]{karunratanakul2023guided}
Korrawe Karunratanakul, Konpat Preechakul, Supasorn Suwajanakorn, and Siyu Tang.
\newblock Guided motion diffusion for controllable human motion synthesis.
\newblock In \emph{Proceedings of the IEEE/CVF International Conference on Computer Vision}, pages 2151--2162, 2023.

\bibitem[Li et~al.(2023)Li, He, Wang, Li, Wang, Luo, Wang, Wang, and Qiao]{2023videochat}
Kunchang Li, Yinan He, Yi Wang, Yizhuo Li, Wenhai Wang, Ping Luo, Yali Wang, Limin Wang, and Yu Qiao.
\newblock Videochat: Chat-centric video understanding.
\newblock \emph{arXiv preprint arXiv:2305.06355}, 2023.

\bibitem[Lin(2004)]{lin2004rouge}
Chin-Yew Lin.
\newblock Rouge: A package for automatic evaluation of summaries.
\newblock In \emph{Text summarization branches out}, pages 74--81, 2004.

\bibitem[Liu et~al.(2023)Liu, Jiang, Zhang, Liu, Zhang, Biswas, and Stone]{liu2023llm+}
Bo Liu, Yuqian Jiang, Xiaohan Zhang, Qiang Liu, Shiqi Zhang, Joydeep Biswas, and Peter Stone.
\newblock Llm+ p: Empowering large language models with optimal planning proficiency.
\newblock \emph{arXiv preprint arXiv:2304.11477}, 2023.

\bibitem[OpenAI.(2023)]{chatgpt}
OpenAI.
\newblock Chatgpt. accessed: 2023-05-15.
\newblock \emph{https://openai.com/blog/chatgpt}, 2023.

\bibitem[Ouyang et~al.(2022{\natexlab{a}})Ouyang, Wu, Jiang, Almeida, Wainwright, Mishkin, Zhang, Agarwal, Slama, Ray, Schulman, Hilton, Kelton, Miller, Simens, Askell, Welinder, Christiano, Leike, and Lowe]{instructGPT22}
Long Ouyang, Jeff Wu, Xu Jiang, Diogo Almeida, Carroll~L. Wainwright, Pamela Mishkin, Chong Zhang, Sandhini Agarwal, Katarina Slama, Alex Ray, John Schulman, Jacob Hilton, Fraser Kelton, Luke Miller, Maddie Simens, Amanda Askell, Peter Welinder, Paul Christiano, Jan Leike, and Ryan Lowe.
\newblock Training language models to follow instructions with human feedback, 2022{\natexlab{a}}.

\bibitem[Ouyang et~al.(2022{\natexlab{b}})Ouyang, Wu, Jiang, Almeida, Wainwright, Mishkin, Zhang, Agarwal, Slama, Ray, et~al.]{ouyang2022training}
Long Ouyang, Jeff Wu, Xu Jiang, Diogo Almeida, Carroll~L Wainwright, Pamela Mishkin, Chong Zhang, Sandhini Agarwal, Katarina Slama, Alex Ray, et~al.
\newblock Training language models to follow instructions with human feedback, 2022.
\newblock \emph{URL https://arxiv. org/abs/2203.02155}, 13, 2022{\natexlab{b}}.

\bibitem[Papineni et~al.(2002)Papineni, Roukos, Ward, and Zhu]{papineni2002bleu}
Kishore Papineni, Salim Roukos, Todd Ward, and Wei-Jing Zhu.
\newblock Bleu: a method for automatic evaluation of machine translation.
\newblock In \emph{Proceedings of the 40th annual meeting of the Association for Computational Linguistics}, pages 311--318, 2002.

\bibitem[Park et~al.(2023)Park, O'Brien, Cai, Morris, Liang, and Bernstein]{park2023generative}
Joon~Sung Park, Joseph~C O'Brien, Carrie~J Cai, Meredith~Ringel Morris, Percy Liang, and Michael~S Bernstein.
\newblock Generative agents: Interactive simulacra of human behavior.
\newblock \emph{arXiv preprint arXiv:2304.03442}, 2023.

\bibitem[Petrovich et~al.(2021)Petrovich, Black, and Varol]{petrovich2021action}
Mathis Petrovich, Michael~J Black, and G{\"u}l Varol.
\newblock Action-conditioned 3d human motion synthesis with transformer vae.
\newblock In \emph{Proceedings of the IEEE/CVF International Conference on Computer Vision}, pages 10985--10995, 2021.

\bibitem[Petrovich et~al.(2022)Petrovich, Black, and Varol]{petrovich2022temos}
Mathis Petrovich, Michael~J Black, and G{\"u}l Varol.
\newblock Temos: Generating diverse human motions from textual descriptions.
\newblock \emph{arXiv preprint arXiv:2204.14109}, 2022.

\bibitem[Petrovich et~al.(2023)Petrovich, Black, and Varol]{petrovich2023tmr}
Mathis Petrovich, Michael~J Black, and G{\"u}l Varol.
\newblock Tmr: Text-to-motion retrieval using contrastive 3d human motion synthesis.
\newblock \emph{arXiv preprint arXiv:2305.00976}, 2023.

\bibitem[Plappert et~al.(2018)Plappert, Mandery, and Asfour]{plappert2018learning}
Matthias Plappert, Christian Mandery, and Tamim Asfour.
\newblock Learning a bidirectional mapping between human whole-body motion and natural language using deep recurrent neural networks.
\newblock \emph{Robotics and Autonomous Systems}, 109:\penalty0 13--26, 2018.

\bibitem[Radford et~al.(2018)Radford, Narasimhan, Salimans, Sutskever, et~al.]{radford2018improving}
Alec Radford, Karthik Narasimhan, Tim Salimans, Ilya Sutskever, et~al.
\newblock Improving language understanding by generative pre-training.
\newblock 2018.

\bibitem[Radford et~al.(2019)Radford, Wu, Child, Luan, Amodei, Sutskever, et~al.]{radford2019language}
Alec Radford, Jeffrey Wu, Rewon Child, David Luan, Dario Amodei, Ilya Sutskever, et~al.
\newblock Language models are unsupervised multitask learners.
\newblock \emph{OpenAI blog}, 1\penalty0 (8):\penalty0 9, 2019.

\bibitem[Raffel et~al.(2020)Raffel, Shazeer, Roberts, Lee, Narang, Matena, Zhou, Li, and Liu]{raffel2020exploring}
Colin Raffel, Noam Shazeer, Adam Roberts, Katherine Lee, Sharan Narang, Michael Matena, Yanqi Zhou, Wei Li, and Peter~J Liu.
\newblock Exploring the limits of transfer learning with a unified text-to-text transformer.
\newblock \emph{The Journal of Machine Learning Research}, 21\penalty0 (1):\penalty0 5485--5551, 2020.

\bibitem[Reed et~al.(2022)Reed, Zolna, Parisotto, Colmenarejo, Novikov, Barth-Maron, Gimenez, Sulsky, Kay, Springenberg, Eccles, Bruce, Razavi, Edwards, Heess, Chen, Hadsell, Vinyals, Bordbar, and de~Freitas]{reed2022generalist}
Scott Reed, Konrad Zolna, Emilio Parisotto, Sergio~Gomez Colmenarejo, Alexander Novikov, Gabriel Barth-Maron, Mai Gimenez, Yury Sulsky, Jackie Kay, Jost~Tobias Springenberg, Tom Eccles, Jake Bruce, Ali Razavi, Ashley Edwards, Nicolas Heess, Yutian Chen, Raia Hadsell, Oriol Vinyals, Mahyar Bordbar, and Nando de Freitas.
\newblock A generalist agent, 2022.

\bibitem[Robertson(2004)]{robertson2004understanding}
Stephen Robertson.
\newblock Understanding inverse document frequency: on theoretical arguments for idf.
\newblock \emph{Journal of documentation}, 60\penalty0 (5):\penalty0 503--520, 2004.

\bibitem[Shafir et~al.(2023)Shafir, Tevet, Kapon, and Bermano]{shafir2023human}
Yonatan Shafir, Guy Tevet, Roy Kapon, and Amit~H Bermano.
\newblock Human motion diffusion as a generative prior.
\newblock \emph{arXiv preprint arXiv:2303.01418}, 2023.

\bibitem[Song et~al.(2023)Song, Wu, Washington, Sadler, Chao, and Su]{song2023llm}
Chan~Hee Song, Jiaman Wu, Clayton Washington, Brian~M Sadler, Wei-Lun Chao, and Yu Su.
\newblock Llm-planner: Few-shot grounded planning for embodied agents with large language models.
\newblock In \emph{Proceedings of the IEEE/CVF International Conference on Computer Vision}, pages 2998--3009, 2023.

\bibitem[Sun et~al.(2023)Sun, Zhuang, Kong, Dai, and Zhang]{sun2023adaplanner}
Haotian Sun, Yuchen Zhuang, Lingkai Kong, Bo Dai, and Chao Zhang.
\newblock Adaplanner: Adaptive planning from feedback with language models, 2023.

\bibitem[Tevet et~al.(2022)Tevet, Raab, Gordon, Shafir, Cohen-Or, and Bermano]{tevet2022human}
Guy Tevet, Sigal Raab, Brian Gordon, Yonatan Shafir, Daniel Cohen-Or, and Amit~H Bermano.
\newblock Human motion diffusion model.
\newblock \emph{arXiv preprint arXiv:2209.14916}, 2022.

\bibitem[Touvron et~al.(2023{\natexlab{a}})Touvron, Lavril, Izacard, Martinet, Lachaux, Lacroix, Rozi{\`e}re, Goyal, Hambro, Azhar, et~al.]{touvron2023llama}
Hugo Touvron, Thibaut Lavril, Gautier Izacard, Xavier Martinet, Marie-Anne Lachaux, Timoth{\'e}e Lacroix, Baptiste Rozi{\`e}re, Naman Goyal, Eric Hambro, Faisal Azhar, et~al.
\newblock Llama: Open and efficient foundation language models.
\newblock \emph{arXiv preprint arXiv:2302.13971}, 2023{\natexlab{a}}.

\bibitem[Touvron et~al.(2023{\natexlab{b}})Touvron, Martin, Stone, Albert, Almahairi, Babaei, Bashlykov, Batra, Bhargava, Bhosale, et~al.]{touvron2023llama2}
Hugo Touvron, Louis Martin, Kevin Stone, Peter Albert, Amjad Almahairi, Yasmine Babaei, Nikolay Bashlykov, Soumya Batra, Prajjwal Bhargava, Shruti Bhosale, et~al.
\newblock Llama 2: Open foundation and fine-tuned chat models.
\newblock \emph{arXiv preprint arXiv:2307.09288}, 2023{\natexlab{b}}.

\bibitem[Vedantam et~al.(2015)Vedantam, Lawrence~Zitnick, and Parikh]{vedantam2015cider}
Ramakrishna Vedantam, C Lawrence~Zitnick, and Devi Parikh.
\newblock Cider: Consensus-based image description evaluation.
\newblock In \emph{Proceedings of the IEEE conference on computer vision and pattern recognition}, pages 4566--4575, 2015.

\bibitem[Wei et~al.(2022)Wei, Bosma, Zhao, Guu, Yu, Lester, Du, Dai, and Le]{wei2022finetuned}
Jason Wei, Maarten Bosma, Vincent~Y. Zhao, Kelvin Guu, Adams~Wei Yu, Brian Lester, Nan Du, Andrew~M. Dai, and Quoc~V. Le.
\newblock Finetuned language models are zero-shot learners, 2022.

\bibitem[Xiao et~al.(2023)Xiao, Wang, Wang, Cao, Zhang, Dai, Lin, and Pang]{xiao2023unified}
Zeqi Xiao, Tai Wang, Jingbo Wang, Jinkun Cao, Wenwei Zhang, Bo Dai, Dahua Lin, and Jiangmiao Pang.
\newblock Unified human-scene interaction via prompted chain-of-contacts.
\newblock \emph{arXiv preprint arXiv:2309.07918}, 2023.

\bibitem[Yuan et~al.(2023)Yuan, Song, Iqbal, Vahdat, and Kautz]{yuan2023physdiff}
Ye Yuan, Jiaming Song, Umar Iqbal, Arash Vahdat, and Jan Kautz.
\newblock Physdiff: Physics-guided human motion diffusion model.
\newblock In \emph{Proceedings of the IEEE/CVF International Conference on Computer Vision}, pages 16010--16021, 2023.

\bibitem[Zhang et~al.(2023{\natexlab{a}})Zhang, Zhang, Cun, Huang, Zhang, Zhao, Lu, and Shen]{zhang2023t2m}
Jianrong Zhang, Yangsong Zhang, Xiaodong Cun, Shaoli Huang, Yong Zhang, Hongwei Zhao, Hongtao Lu, and Xi Shen.
\newblock T2m-gpt: Generating human motion from textual descriptions with discrete representations.
\newblock \emph{arXiv preprint arXiv:2301.06052}, 2023{\natexlab{a}}.

\bibitem[Zhang et~al.(2023{\natexlab{b}})Zhang, Guo, Pan, Cai, Hong, Li, Yang, and Liu]{zhang2023remodiffuse}
Mingyuan Zhang, Xinying Guo, Liang Pan, Zhongang Cai, Fangzhou Hong, Huirong Li, Lei Yang, and Ziwei Liu.
\newblock Remodiffuse: Retrieval-augmented motion diffusion model.
\newblock \emph{arXiv preprint arXiv:2304.01116}, 2023{\natexlab{b}}.

\bibitem[Zhang et~al.(2019)Zhang, Kishore, Wu, Weinberger, and Artzi]{zhang2019bertscore}
Tianyi Zhang, Varsha Kishore, Felix Wu, Kilian~Q Weinberger, and Yoav Artzi.
\newblock Bertscore: Evaluating text generation with bert.
\newblock \emph{arXiv preprint arXiv:1904.09675}, 2019.

\bibitem[Zhang et~al.(2023{\natexlab{c}})Zhang, Huang, Liu, Tang, Lu, Chen, Bai, Chu, Yu, and Ouyang]{zhang2023motiongpt}
Yaqi Zhang, Di Huang, Bin Liu, Shixiang Tang, Yan Lu, Lu Chen, Lei Bai, Qi Chu, Nenghai Yu, and Wanli Ouyang.
\newblock Motiongpt: Finetuned llms are general-purpose motion generators.
\newblock \emph{arXiv preprint arXiv:2306.10900}, 2023{\natexlab{c}}.

\bibitem[Zhou and Wang(2023)]{zhou2023ude}
Zixiang Zhou and Baoyuan Wang.
\newblock Ude: A unified driving engine for human motion generation.
\newblock In \emph{Proceedings of the IEEE/CVF Conference on Computer Vision and Pattern Recognition}, pages 5632--5641, 2023.

\bibitem[Zhu et~al.(2023)Zhu, Ma, Liu, Liu, Wu, and Wang]{zhu2023motionbert}
Wentao Zhu, Xiaoxuan Ma, Zhaoyang Liu, Libin Liu, Wayne Wu, and Yizhou Wang.
\newblock Motionbert: A unified perspective on learning human motion representations.
\newblock In \emph{Proceedings of the IEEE/CVF International Conference on Computer Vision}, pages 15085--15099, 2023.

\end{thebibliography}
}

\clearpage
\appendix

\twocolumn[
\begin{@twocolumnfalse}
\section*{\centering{Supplementary Material}}
\end{@twocolumnfalse}
]

\section{Prompts for Instruction Tuning}We define seven major motion-related tasks, including 1) motion generation(MG), 2) motion understanding(MU), 3) motion-in-between(MiB), 4) task planning(CT2T), 5) task decomposition(T2S), 6) task summarization(S2T), and 7) scene estimation(T2C). We also define another four high-level tasks as auxiliary in training to assist in learning the comprehensive relationship between motion and language descriptions. These auxiliary tasks are: 1) steps planning conditioned on historical task(CT2S), 2) task planning conditioned on historical steps(CS2T), 3) steps planning conditioned on historical steps(CS2S), 4) scene estimation from steps(S2C). The notation of `C', `S', and `T' are identical as previously stated.

\begin{figure*}
    \centering
    \includegraphics[width=1.0\linewidth]{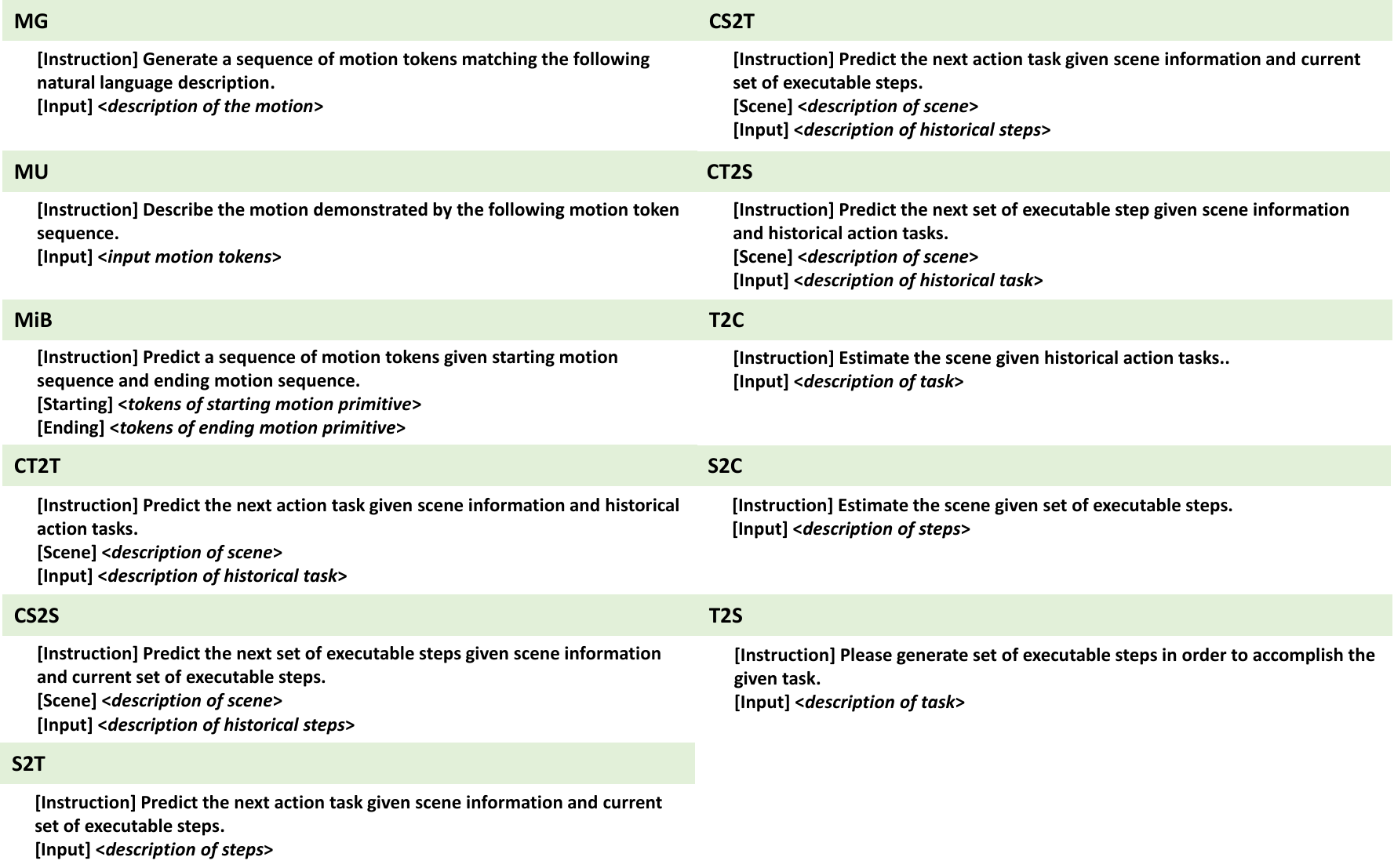}
    \caption{\textbf{Prompts for Instruction Tuning.} We present the prompts we adopted for instruction tuning our model. \textit{$\langle text \rangle$} indicates the input text/motion tokens and output text/motion tokens.}
    \label{fig:supp-instruction-tunning}
\end{figure*}

The prompts employed for instruction tuning the shared LLM on these tasks are shown in Fig. \ref{fig:supp-instruction-tunning}.

\section{Automatic Annotation Pipeline}The proposed annotation pipeline contains five modules covering video preprocessing and video describing at different levels of detail. 

\paragraph{Video Processing}The raw input videos could be up to hours long, which contain rich content. It is not practical to describe the content of the entire video at once. Therefore, we perform the video cropping to break down it into multiple short video segments. Experiments show that breaking down into 10-second chunks balances between efficiency and annotation richness.

\begin{figure}
    \centering
    \includegraphics[width=1.0\linewidth]{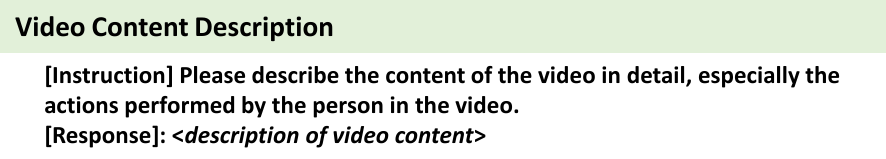}
    \caption{\textbf{Prompts for Video Describing.} We generate content descriptions using this prompt. \textit{$\langle text \rangle$} indicates the output description.}
    \label{fig:supp-vllm}
\end{figure}

\paragraph{Video Content Describing}We use Ask-Anything\cite{2023videochat} to describe the content of each video segment. It generates one long textual description with up to 500 words to depict the person's activity, appearance, and environment shown in the video segment. Fig. \ref{fig:supp-vllm} presents the prompts used for this task.

\paragraph{Motion Describing}We use ChatGPT to extract key information from the video content description. What we are interested in are those describing human motion and activity. The experiment shows that hierarchical extraction obtains results that meet our requirements best. Specifically, we ask ChatGPT to extract information that describes how the person in the video executes the activity step-by-step. It produces several short descriptions at the finest text granularity(\textbf{step description}). Following this, we ask ChatGPT to summarize the contents of these step-by-step descriptions, the results depict the specific activities at medium-grained text granularity(\textbf{task description}). We also generate content description that depicts the scene information of the entire video at coarse-grained granularity(\textbf{scene description}). For this purpose, we feed the task descriptions to ChatGPT, we emphasize their order because temporal sequence matters. 

\begin{figure}
    \centering
    \includegraphics[width=1.0\linewidth]{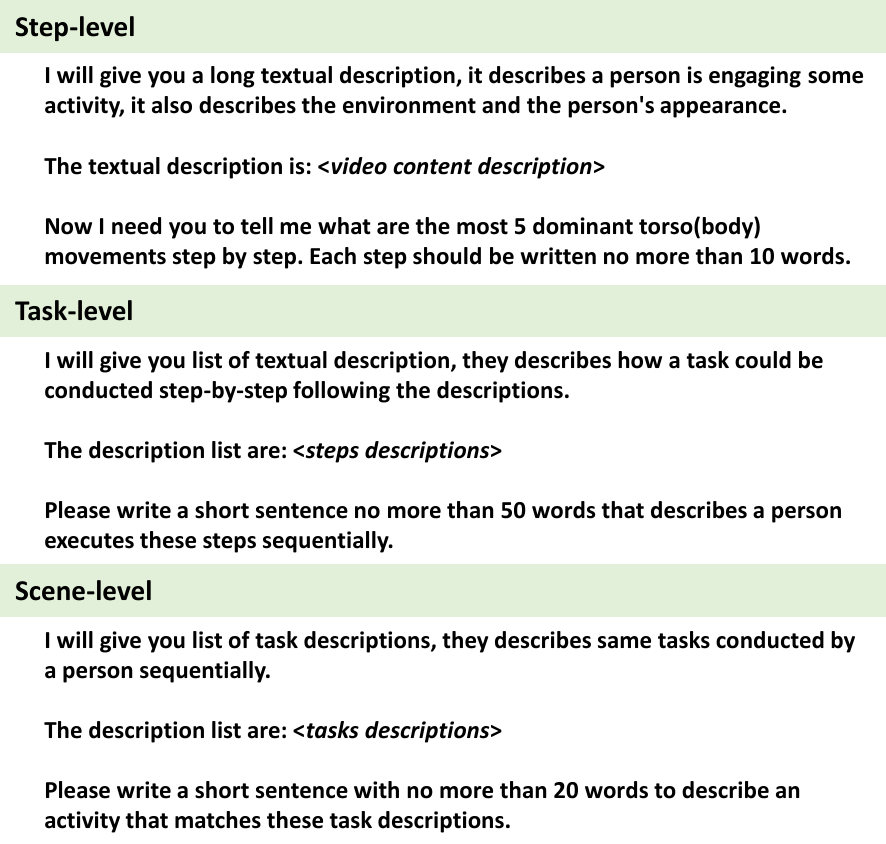}
    \caption{\textbf{Prompts for Motion Describing.} We present the prompts for hierarchical annotation generation. \textit{$\langle text \rangle$} indicates the input/output texts.}
    \label{fig:supp-annotation}
\end{figure}

We use the following prompts shown in Fig. \ref{fig:supp-annotation} to extract step-, task- and scene-level descriptions.

\begin{figure}
    \centering
    \includegraphics[width=1.0\linewidth]{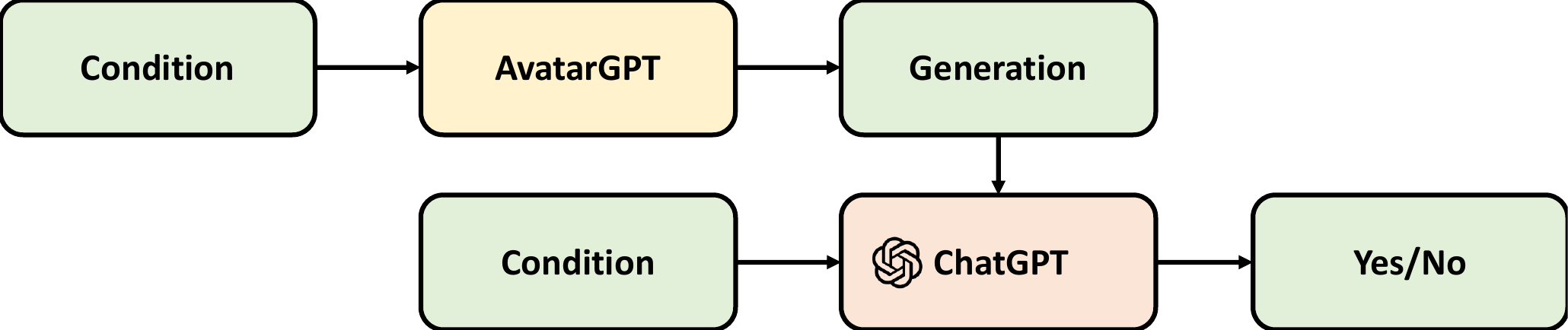}
    \caption{\textbf{Logical Coherency Evaluation.} We ask ChatGPT to determine whether the generation matches the condition in terms of logical coherency.}
    \label{fig:supp-high-metric}
\end{figure}

\section{Metrics for Low-level Tasks}We describe the metrics for linguistic consistency evaluation here.

\paragraph{BertScore\cite{zhang2019bertscore}} uses a pretrained BERT model to evaluate the similarity between generated and reference text. It encodes the reference and target sentence to embeddings and estimates the cosine similarity between embeddings are the similarity score.

\paragraph{BLUE\cite{papineni2002bleu}} measures the sub-sentence level(1-gram, 4-gram) correctness between candidate sentence and set of references. Although it was proposed for translation initially, it is widely adopted in assessing text generation.

\paragraph{ROUGE-L\cite{lin2004rouge}} assesses the similarity between two sentences according to the longest common sub-sequence(LCS).

\paragraph{CIDEr\cite{vedantam2015cider}} measures the similarity based on the concept of consensus in terms of word, grammar, and text content. It first estimates the BLUE score between target and references and modifies the scores using inverse document frequency(IDF) weighting\cite{robertson2004understanding} to balance the weights between rare and common words. Finally, it averages the weighted scores as the final CIDEr score.

\section{Metrics for High-level Tasks} 
\begin{figure*}
    \centering
    \includegraphics[width=1.0\linewidth]{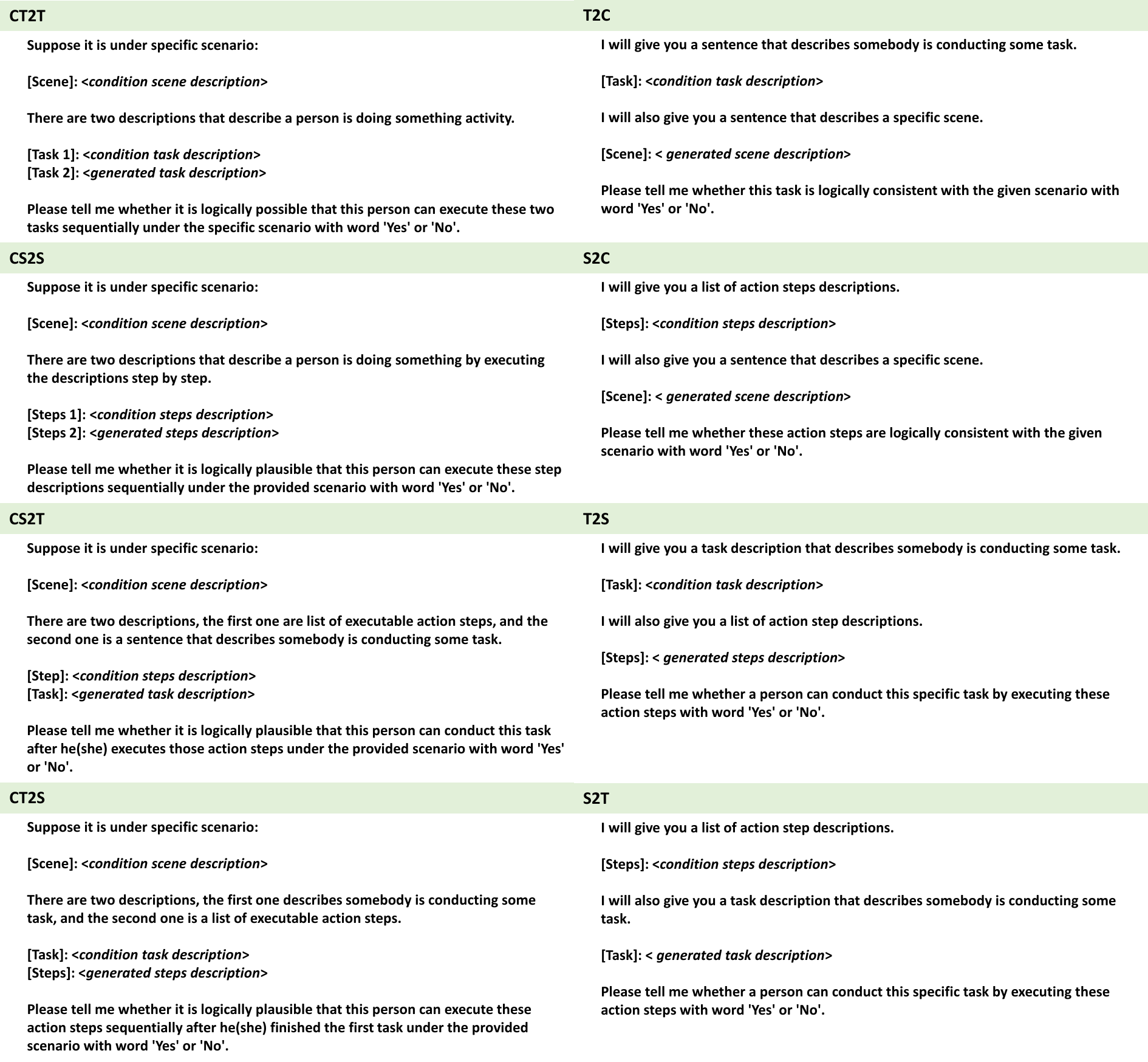}
    \caption{\textbf{Prompts for High-level Tasks Evaluation.} We show prompts we employ to conduct automatic high-level task evaluation. \textit{$\langle text \rangle$} indicates the input/output texts.}
    \label{fig:supp-highlevel-prompt}
\end{figure*}

The logical coherence between generated text and conditions is the primary concern. For instance, if the condition scene is \textit{`indoor fitness exercise'
}, and the historical activity is \textit{`push-ups'}, one logically coherent activity could be \textit{`
jumping jacks'}, while \textit{`organizing items on the ground'} is incoherent. We harness the capability of understanding and reasoning of ChatGPT to evaluate the consistency. Fig. \ref{fig:supp-high-metric} illustrates how ChatGPT is employed to assist the logical coherency evaluation on high-level tasks. Fig. \ref{fig:supp-highlevel-prompt} shows the prompts used for each evaluation task.

\section{Full Pipeline Evaluation}
\begin{figure*}
    \centering
    \includegraphics[width=1.0\linewidth]{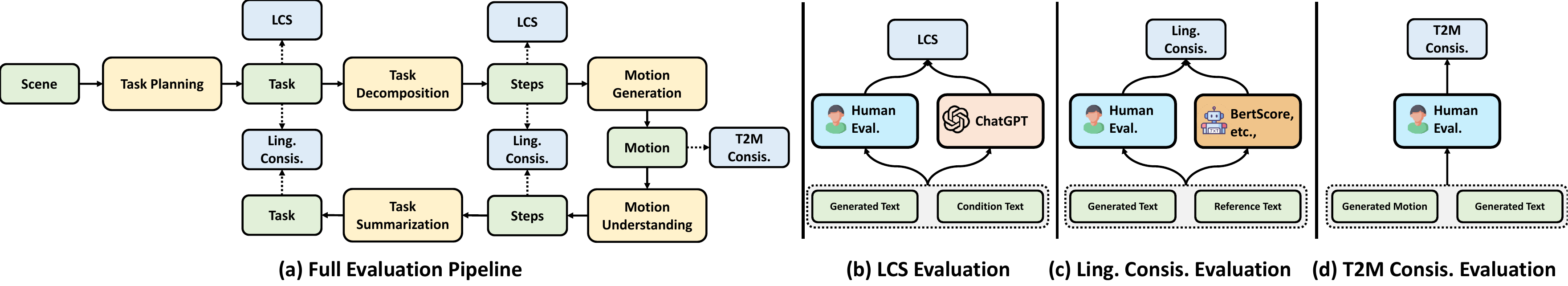}
    \caption{\textbf{Method for Full Pipeline Evaluation.} We measure the performance of our method in the scope of the full pipeline based on the concept of cycle consistency. (a) Evaluation pipeline. (b) Evaluate the logical coherent score(LCS) using human evaluation and ChatGPT. (c) Evaluate the linguistic consistency(Ling. Consis.) using human evaluation and quantitative metrics(e.g. BertScore). (d) Evaluate the text-to-motion consistency(T2M Consis.) using human evaluation.}
    \label{fig:supp-full}
\end{figure*}

The full pipeline of our method includes task planning, decomposition, motion generation, understanding, in-between, etc. We conduct quantitative evaluation and user study to assess the performance. The entire evaluation workflow is shown in Fig. \ref{fig:supp-full}. We define a forward path as scene $\rightarrow$ task $\rightarrow$ steps $\rightarrow$ motions, where task planning, decomposition, motion generation, and in-between are conducted sub-sequentially. The backward path executes the motion understanding and task summarization one after another, resulting in the following outputs: motion $\rightarrow$ steps $\rightarrow$ task. 

We perform quantitative analysis and human evaluation to assess the linguistic consistency(Ling. Consis.) between planned and summarized tasks and decomposed and described step descriptions. Specifically, we estimate the BertScore, BLUE, ROUGE-L, and CIDEr scores as quantitative metrics, and we ask human participants to judge whether the target and reference sentences have similar semantic content. Score `1' is expected if semantic similarity exists, otherwise, `0'. 

In addition, quantitative and human evaluations are also executed to measure the logical coherent score(LCS) in terms of task planning and decomposition in the forward path. For quantitative analysis, the same approach discussed in Appendix D is adopted. For human evaluation, users are expected to score `1' if they believe the generated texts are logically coherent to condition texts, otherwise, a score `0' is given. 

We also conducted a user study to evaluate the consistency between generated long motion sequences and condition descriptions. Because the final motion is the combination of synthesized short motion segments corresponding to multiple condition descriptions, and the blended motions in-between, it is impractical to simply give a yes/no grade. We propose a rating scale of 1-5 for this task. Specifically, users are expected to score `5' if the generated motion presents all contents described in corresponding conditions. On the contrary, score `1' is expected if none of the descriptions are correctly presented in the generated motion. For those motions that present partial conditions, scores between `1' and `5' are expected according to the percentage of conditions synthesized.

\section{Results of Automatic Video Annotation}
\begin{figure*}
    \centering
    \includegraphics[width=0.8\linewidth]{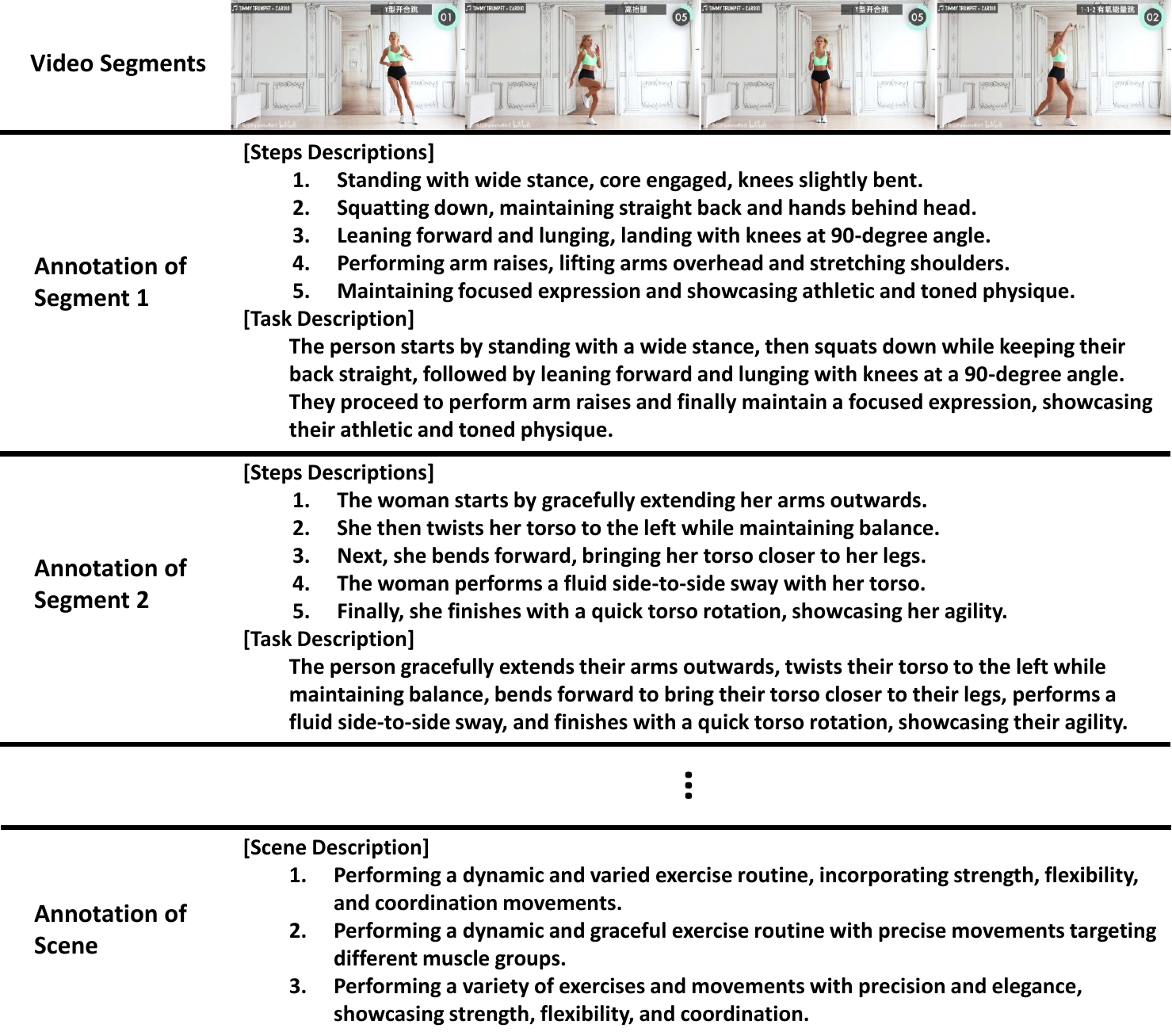}
    \caption{\textbf{Annotation Results.} We show one task description and five-step descriptions for each video segment. We show three scene descriptions of the entire video.}
    \label{fig:supp-annotation-result}
\end{figure*}

We show one example of video annotation results in Fig. \ref{fig:supp-annotation-result}. Our annotation pipeline crops the input video into segments. We run multiple rounds on each segment to generate descriptions of diverse expressions. Specifically, we run six rounds per segment, resulting in 6 task descriptions and 30 step descriptions. We run twenty rounds on the entire video(task descriptions of all the video segments) to obtain multiple scene descriptions. For brevity, we show the results of two randomly selected segments, each with one task and five-step descriptions. We also show three scene descriptions.

\section{More Results of Motion Generation, Understanding, and In-Between}
\begin{figure*}
    \centering
    \includegraphics[width=0.9\linewidth]{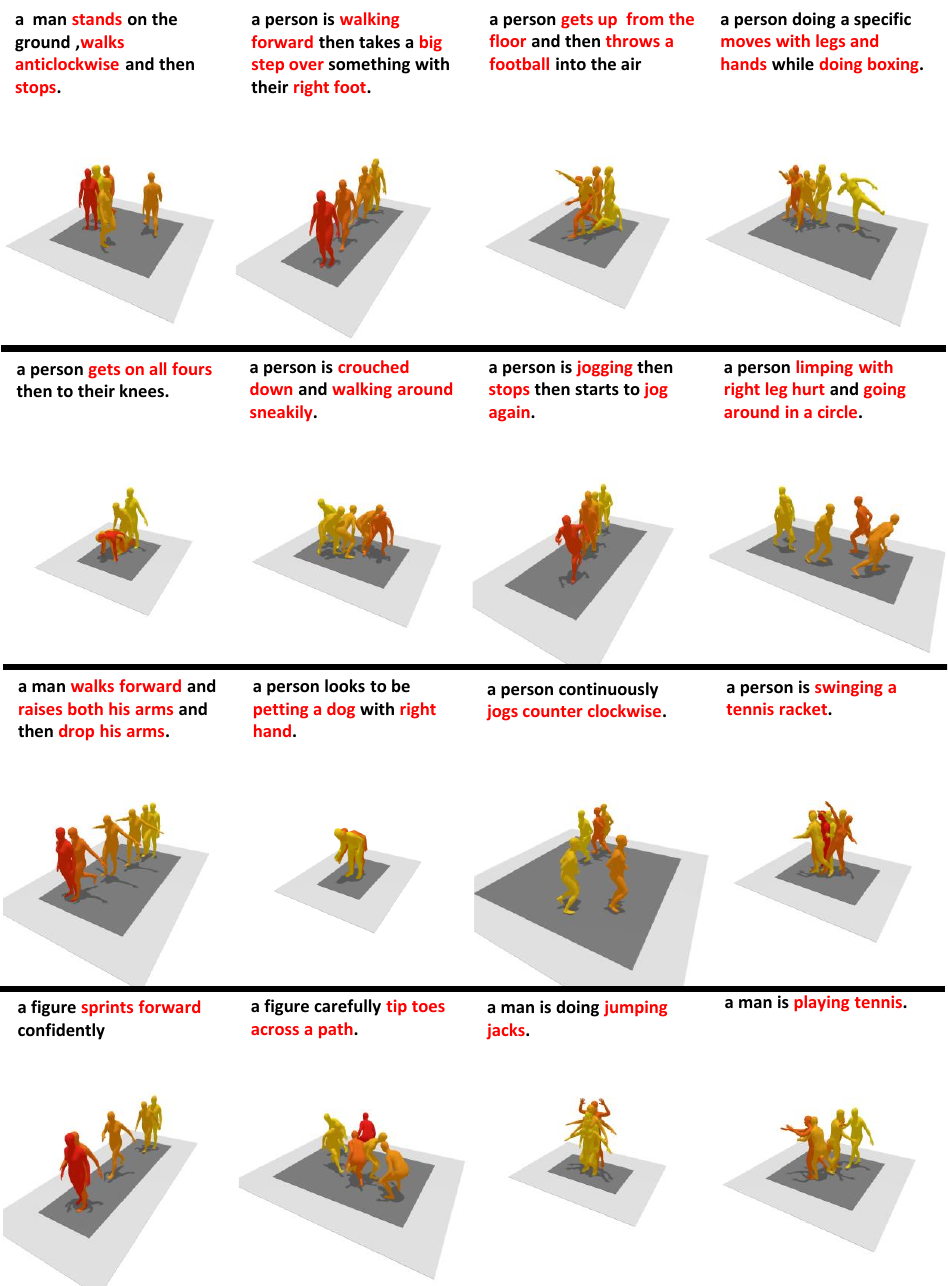}
    \caption{\textbf{Results of Motion Generation.}}
    \label{fig:supp-t2m}
\end{figure*}

\begin{figure*}
    \centering
    \includegraphics[width=0.85\linewidth]{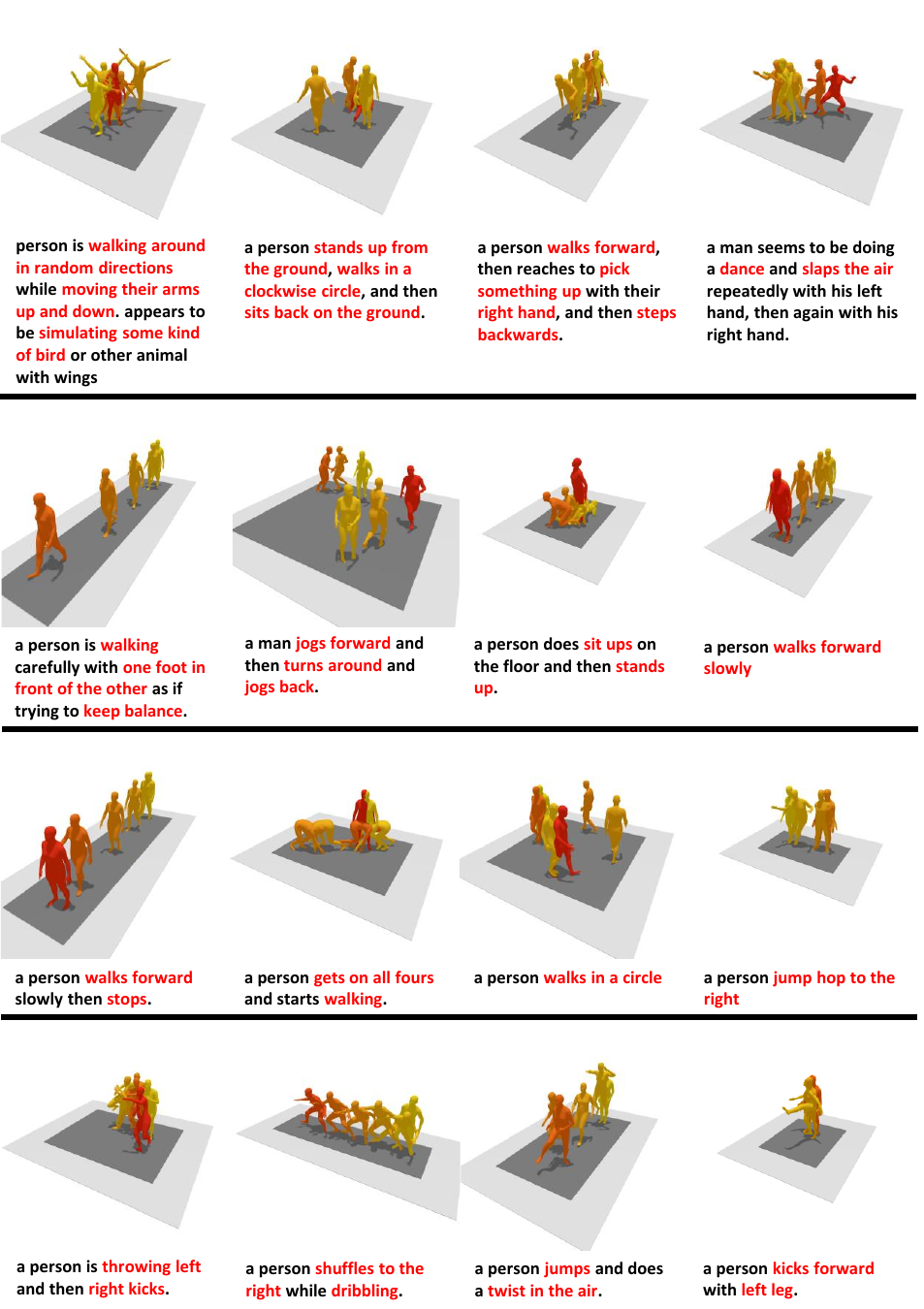}
    \caption{\textbf{Results of Motion Understanding.}}
    \label{fig:supp-m2t}
\end{figure*}

\begin{figure*}
    \centering
    \includegraphics[width=0.9\linewidth]{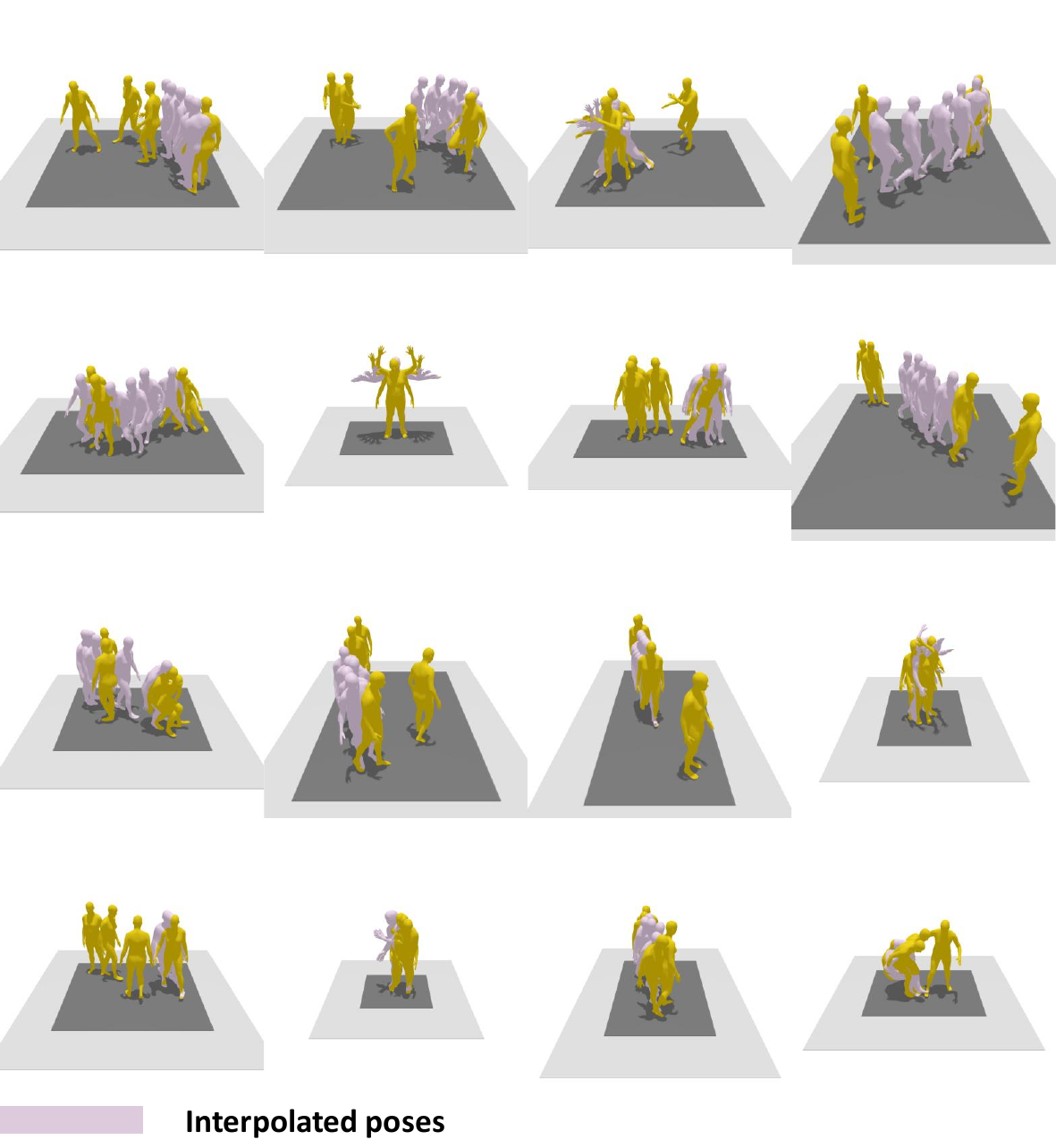}
    \caption{\textbf{Results of Motion-in-Between.}}
    \label{fig:supp-mib}
\end{figure*}

We show more results of motion generation, motion understanding, and motion-in-between tasks in Fig. \ref{fig:supp-t2m}, \ref{fig:supp-m2t} \ref{fig:supp-mib}. For motion generation and understanding tasks, we highlight the key words to emphasize the semantics. For motion-in-between, we use different colors for interpolated frames.

\section{Motion Generation with Various Text Granularity}
\begin{figure*}
    \centering
    \includegraphics[width=0.9\linewidth]{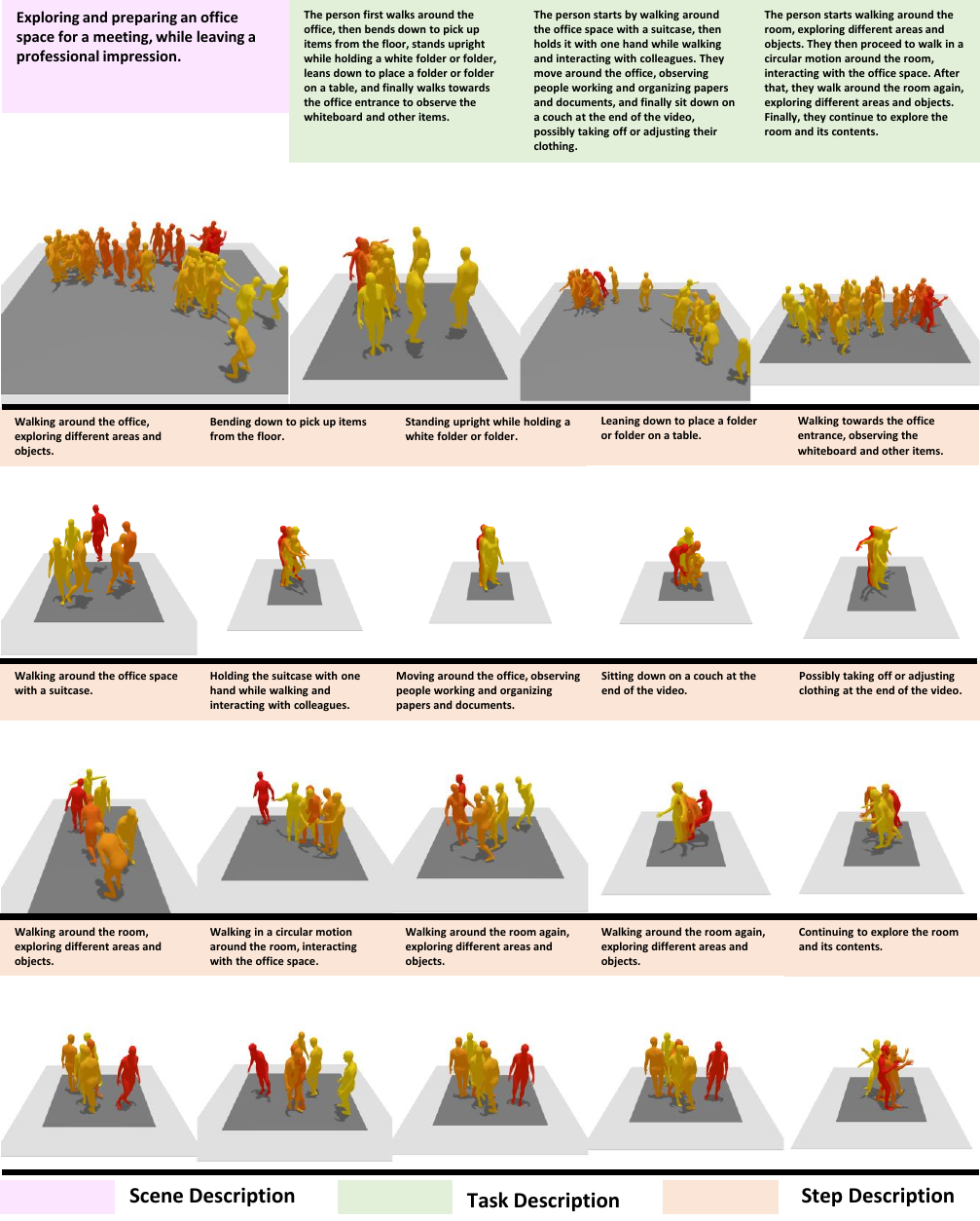}
    \caption{\textbf{Motion Generation with Various Text Granularity.} Our method supports synthesizing human motion from various text granularity. Given a scene description(coarse-grained), our model generates a long motion sequence that matches the context. Given a task description(medium-grained), our method can synthesize a motion sequence that displays the corresponding activity. Given a step description(fine-grained), our method can generate a motion sequence that corresponds to the specific action.}
    \label{fig:supp-t2m-1}
\end{figure*}

\begin{figure*}
    \centering
    \includegraphics[width=0.9\linewidth]{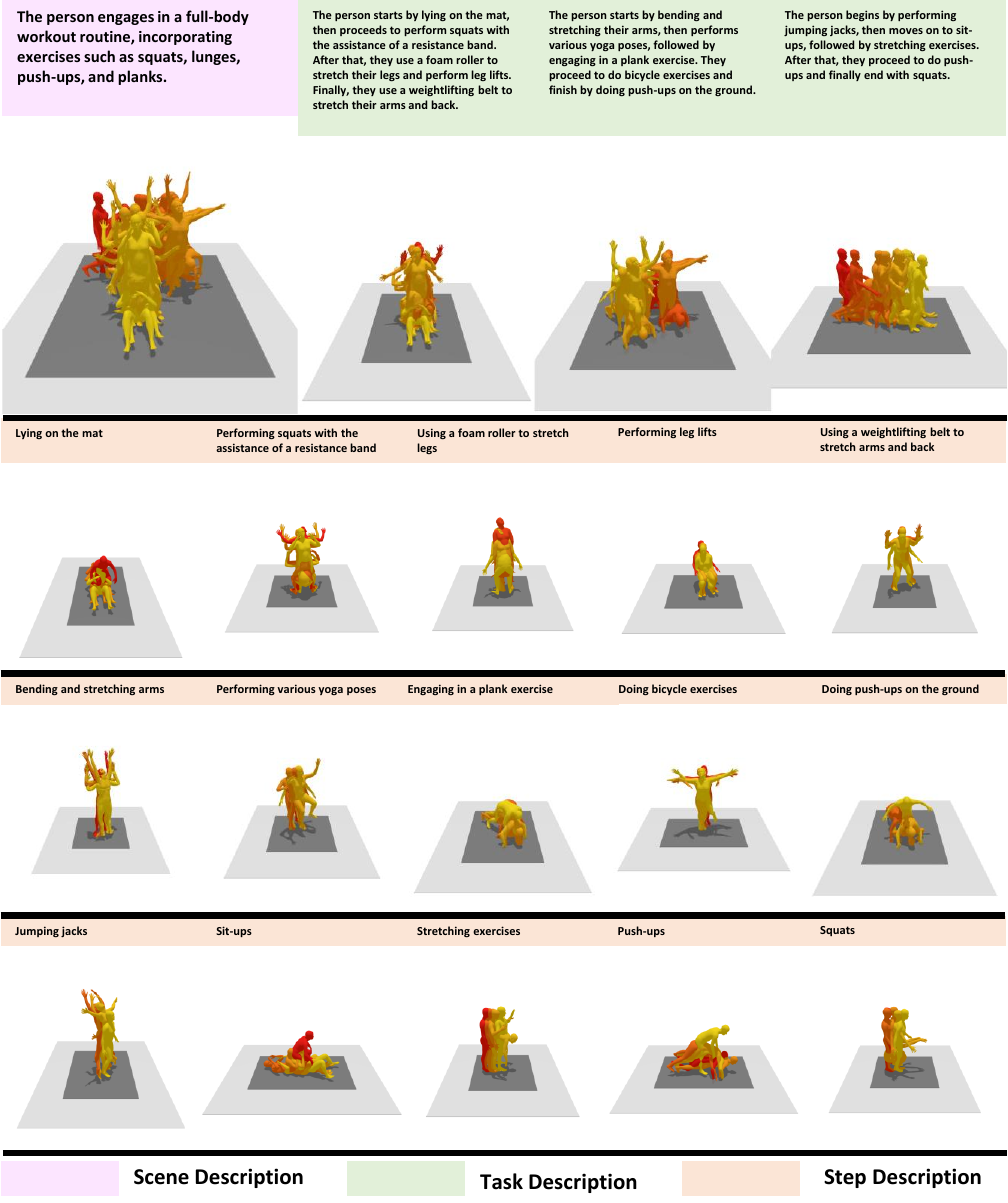}
    \caption{\textbf{Motion Generation with Various Text Granularity.} Our method supports synthesizing human motion from various text granularity. Given a scene description(coarse-grained), our model generates a long motion sequence that matches the context. Given a task description(medium-grained), our method can synthesize a motion sequence that displays the corresponding activity. Given a step description(fine-grained), our method can generate a motion sequence that corresponds to the specific action.}
    \label{fig:supp-t2m-2}
\end{figure*}

We show more results on generating motion sequences from descriptions of various granularity in Fig. \ref{fig:supp-t2m-1}, \ref{fig:supp-t2m-2}. The scene of Fig. \ref{fig:supp-t2m-1} is some regular activities performed in an office space, and that of Fig. \ref{fig:supp-t2m-2} is an indoor fitness workout. We show the motion synthesis conditioned on scene description(coarse-grained), task descriptions(medium-grained), and step descriptions(fine-grained), respectively. We observe high consistency between synthesized motion and conditions, regardless of their text granularities.

\section{Motion Understanding of Various Levels of Detail}
\begin{figure*}
    \centering
    \includegraphics[width=0.75\linewidth]{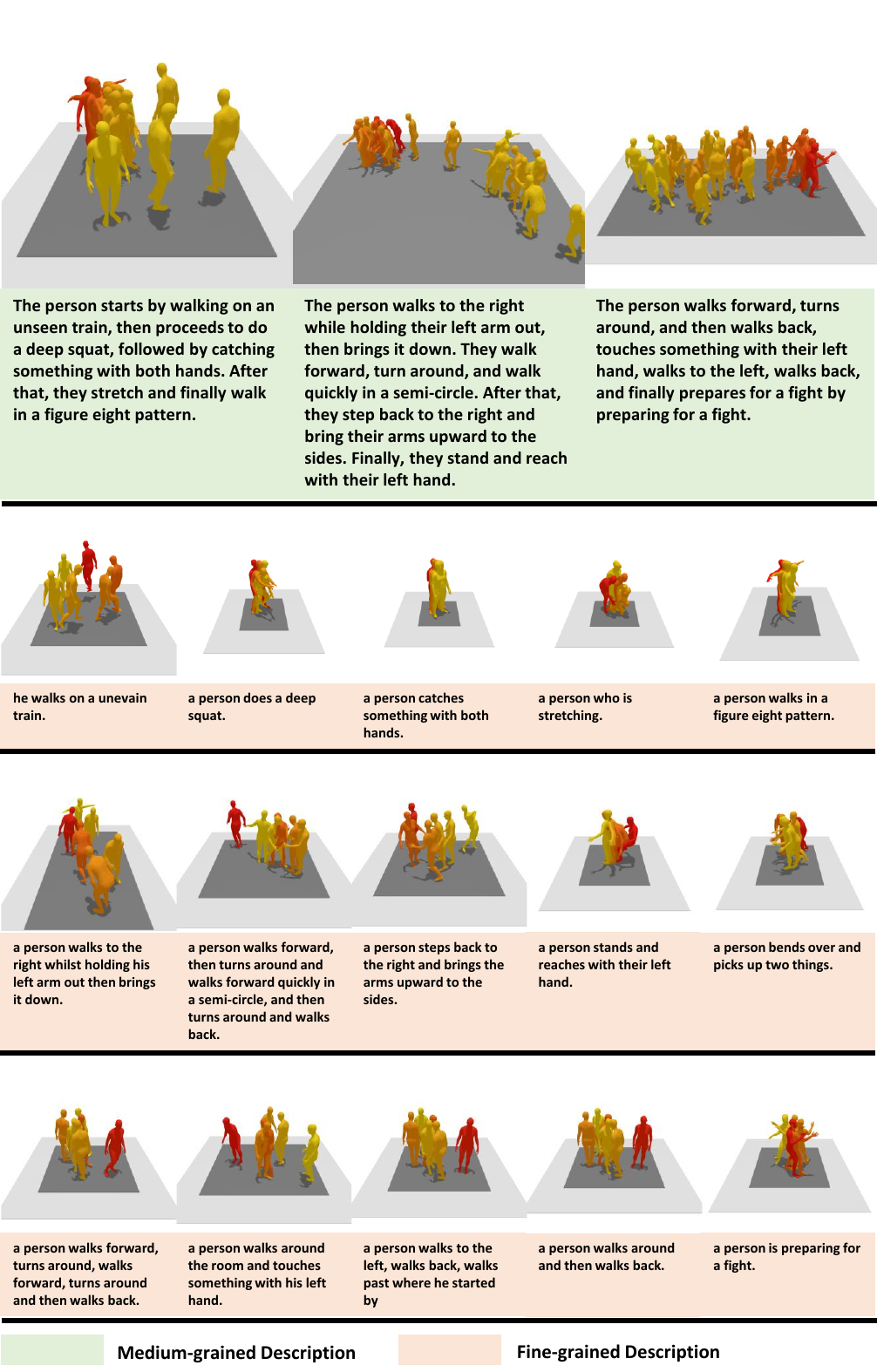}
    \caption{\textbf{Motion Understanding at Various Text Granularity.} Our method can describe motion at various levels of detail. Given long motions, our method can generate descriptions that depict the presented activity(medium-grained). Given a short motion sequence, our method can describe its specific action in a short sentence(fine-grained).}
    \label{fig:supp-m2t-1}
\end{figure*}

\begin{figure*}
    \centering
    \includegraphics[width=0.75\linewidth]{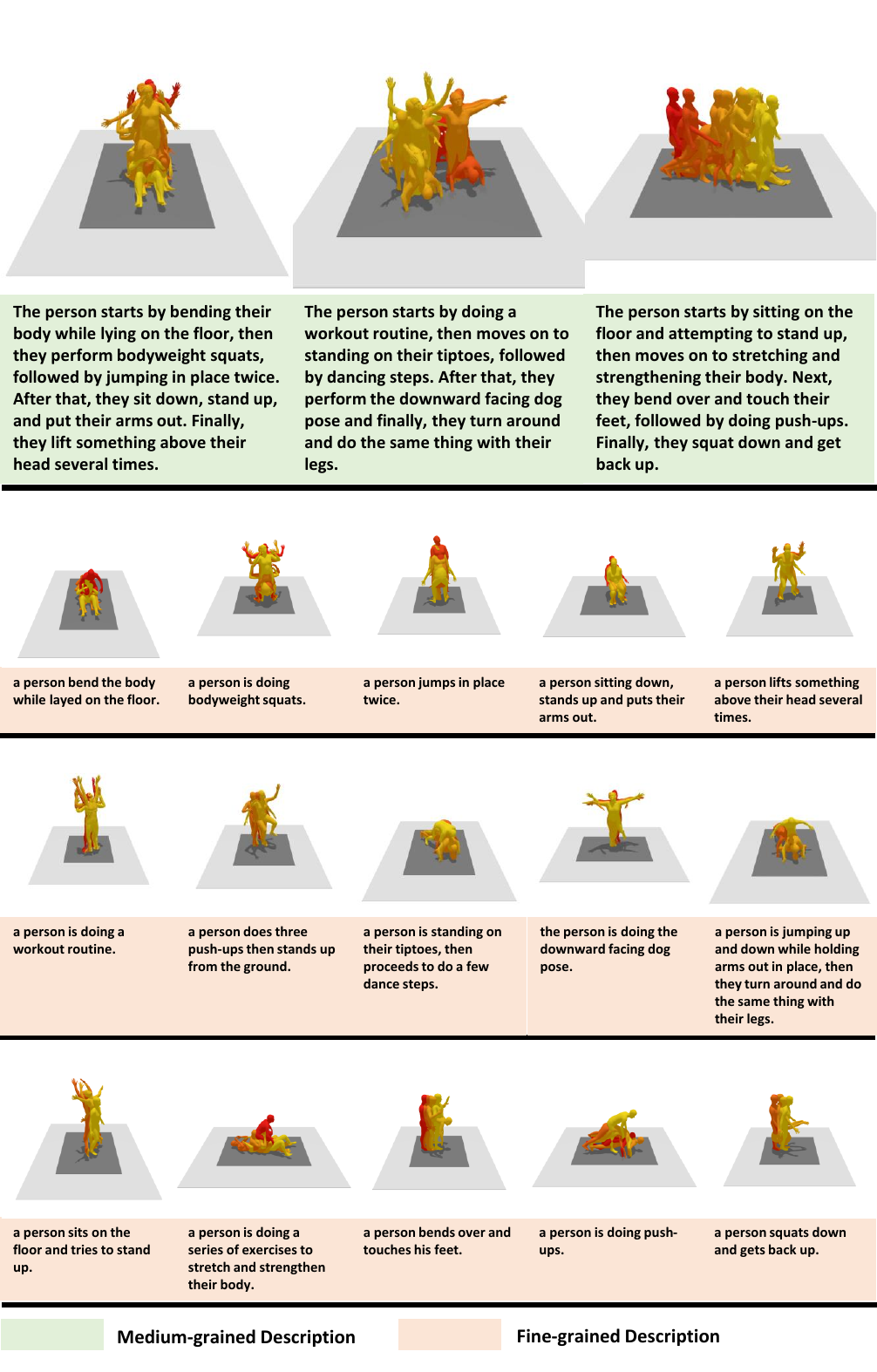}
    \caption{\textbf{Motion Understanding at Various Text Granularity.} Our method can describe motion at various levels of detail. Given long motions, our method can generate descriptions that depict the presented activity(medium-grained). Given a short motion sequence, our method can describe its specific action in a short sentence(fine-grained).}
    \label{fig:supp-m2t-2}
\end{figure*}

We show more results in describing motion sequences at various levels of detail in Fig. \ref{fig:supp-m2t-1}, \ref{fig:supp-m2t-2}. We first show the results of describing the activity displayed by long motion sequences(500+ frames) at medium-grained granularity. Then we present the results of describing the specific actions demonstrated in the short motion segments($\approx$200 frames) at fine-grained granularity. These results suggest that our method has a great ability to understand human movements at various levels of detail.

\section{Limitation and Future Work}1) The training of high-level tasks and low-level tasks are not strictly joint. There is a domain gap between the datasets for low-level and high-level training. Building a larger dataset covering both low- and high-level is one of the top priorities of our future research. 2) Our method is only able to conduct torso movement-related tasks. We believe that facial expression-related tasks are of equal importance. For instance, generating head poses and facial expression sequences subject to the conversation is a good complementary to the torso and gesture movement synthesis. We argue these tasks have the potential to be integrated into one shared pipeline. We will investigate the unification of body, head movements, facial expression, and related high-level tasks in the future. 

\end{document}